\pdfoutput=1
\documentclass[11pt]{article}

\usepackage{EMNLP2022}

\usepackage{times}
\usepackage{latexsym}

\usepackage[T1]{fontenc}
\usepackage[utf8]{inputenc}

\usepackage{microtype}

\usepackage{inconsolata}

\usepackage{booktabs}
\usepackage{enumitem}
\usepackage{fdsymbol}
\usepackage{graphicx}
\usepackage{mathtools}
\usepackage{microtype}
\usepackage{multirow}
\usepackage{subfigure}
\usepackage{xspace}

\newcommand{\our}{\mbox{\textsc{ConDA}}\xspace}
\newcommand{\smallsection}[1]{\noindent\textbf{#1.}}

\newcommand\blfootnote[1]{%
  \begingroup
  \renewcommand\thefootnote{}\footnote{#1}%
  \addtocounter{footnote}{-1}%
  \endgroup
}

\title{Leveraging QA Datasets to Improve Generative Data Augmentation}

\author{
  Dheeraj Mekala$^{\diamondsuit}$ $\qquad$ Tu Vu$^{\spadesuit}$ $\qquad$ Timo Schick$^{\clubsuit}$ $\qquad$  Jingbo Shang$^{\diamondsuit,\heartsuit, *}$ \\
  $^\diamondsuit$ University of California San Diego\\
  $^\spadesuit$ University of Massachusetts Amherst\\
  $^\clubsuit$ Meta AI Research \\
  $^\heartsuit$ Hal\i c\i o\u glu Data Science Institute, University of California San Diego\\
  \small$^{\diamondsuit}$ \texttt{\{dmekala, jshang\}@ucsd.edu} $\qquad$ \small$^{\spadesuit}$ \texttt{tuvu@cs.umass.edu} $\qquad$ \small$^{\clubsuit}$ \texttt{schick@fb.com}
}

\begin{document}
\maketitle

\begin{abstract}
\blfootnote{$*$ Jingbo Shang is the corresponding author.}
% \tu{Intermediate training on QA in the tittle is somewhat confusing/misleading as it reminds readers of STILTS on QA. How about something like leveraging QA for better generative data augmentation?
% Similar comment for the title of Section 3.}
% \timo{I think our method should have a clear and easy to remember name. We don't introduce a name until section 5.3., which in my view is way too late. I would also replace Sections 2 - 4 with a single section 2 that is named after the method, and then convert the current sections 2 - 4 to section 2.1, 2.2 and 2.3, respectively. Maybe we can brainstorm about this later today?}

% \timo{I have made the abstract quite a bit shorter. Feel free to revert these changes if you like the original version more. In particular, I have removed the first sentence because after all, our work is not really about ''annotating'' existing datasets with LMs, but about creating them from scratch.}

% Manually annotating datasets requires domain experts to read through many documents and carefully label them,
% \tu{Don't use things like "thousands" or "millions" here. It varies across tasks and applications. Sometimes they only need to annotate a few or tens of examples} 
% which is often expensive.
% \tu{don't use "too" in a research paper}
% \tu{It's a bit unclear in the title how we are gonna leverage QA datasets, maybe something like Leveraging Context Generation Models for Better Generative Data Augmentation sound better?}
The ability of generative language models (GLMs) to generate text has improved considerably in the last few years, enabling their use for \emph{generative data augmentation}. 
In this work, we propose \our, 
% \tu{What does it stand for? Does \textsc{ConDA} (Con from Context) sound better}, 
an approach to further improve GLMs' ability to generate synthetic data by reformulating data generation as context generation for a given question-answer (QA) pair and leveraging QA datasets for training context generators.
% \tu{GLM's -> GLMs'}\tu{If you mention something like "and fine-tuning these models on QA datasets to obtain context generators" then it makes more sense to  mention "the fine-tuned context generators" in the following sentence}.
% In this work, we propose to further improve their ability to generate synthetic data by reformulating data generation as context generation for a given question-answer (QA) pair and leveraging QA datasets for training context generators.
% \tu{for training context generators sounds better than for intermediate training}.
% In this work, we propose \our, a generative data augmentation method that further improves GLM's ability to generate synthetic data by reformulating data generation as context generation for a given question-answer (QA) pair and leveraging QA datasets for intermediate training \tu{for training context generators sounds better than for intermediate training}.
% Specifically, we view QA to be more as a format than of a task and train GLMs as context generators for a given question and its respective answer.
% \tu{It's still a task, just say we cast QA as context generation, i.e., a model is fed a,q and aksed to produce c} and train GLMs as context generators for a given question and its respective answer.
Then, we cast downstream tasks into the same question answering format and adapt the fine-tuned context generators to the target task domain.
% \tu{Can remove "Then,", ""}\tu{can remove "fine-tuned"}\tu{Can rewrite this sentence as "To adapt the context generators to a target domain, we cast ..."}
Finally, we use the fine-tuned GLM to generate relevant contexts, which are in turn used as synthetic training data for their corresponding tasks.
We perform extensive experiments on multiple classification datasets and demonstrate substantial improvements in performance for both few- and zero-shot settings.
% \timo{I have removed the sentence about SST-2 performance. Seems to be a very specific finding. Maybe we can have some average number, like ``on average, our approach improves few-shot performance by XX\% compared to the strongest baseline considered''?}
% Remarkably, on the SST-2 dataset, intermediate training on SocialIQA dataset achieves an improvement of 40\% on Macro-F1 score.
% Extensive experiments on multiple sentiment and topic classification datasets demonstrate superior performance of our method in few-shot and zero-shot settings.
% \tu{two many and in this sentence} 
% \tu{Could be helpful to include some numbers.}
% \jingbo{I agree with Tu that we should highlight the (average) improvement. it should be impressive here.}
% \tu{demonstrate that our methods result in}
Our analysis reveals that QA datasets that require high-level reasoning abilities (e.g., abstractive and common-sense QA datasets) tend to give the best boost in performance in both few-shot and zero-shot settings.
% \tu{Can remove "in both few-shot and zero-shot settings"}
% \tu{Any other interesting findings from other ablation/analysis experiments?}
% \tu{Any takeaway messages for the paper as a whole?}

% - Manually annotating datasets is expensive
% - Generative language models capabilities to generate text
% - Generate artificial training data
% - Intermediate training on QA datasets.
% - improves upon self-training
% - better than few-shot baselines
\end{abstract}
 
\section{Introduction}

% Deep learning models have been successfully used to solve many NLP tasks~\cite{devlin-etal-2019-bert, liu2019roberta, Radford2018ImprovingLU, 2020t5}, but they typically require large amount of labeled data which is often expensive to obtain.
% % \tu{"require" -> "typically require", I'd be specific about the amount, e.g., thousands to tens of thousands}
% Therefore, limited-data regimes such as zero-shot~\cite{Pushp2017TrainOT, meng-etal-2020-text, alcoforado2022zeroberto}, few-shot~\cite{gao-etal-2021-making, schick-schutze-2021-just} and weak supervision~\cite{meng2018weakly, mekala2020contextualized, wang-etal-2021-x} 
% % \jingbo{we will need to cite other weak supervision papers too} 
% has recently attracted much attention from researchers.
% \tu{"regime" -> "regimes". I'd add zero-shot and cite more on zero-shot/few-shot than weak supervision}

Recent advances in NLP have substantially improved the capability of pretrained language models to generate high-quality text~\cite{Radford2018ImprovingLU, Radford2019LanguageMA, lewis-etal-2020-bart, NEURIPS2020_1457c0d6}.
% Prior work~\cite{vu-etal-2021-strata, mekala-etal-2021-coarse2fine} leverage them for generating synthetic training data, however, generating long documents has not been previously explored.
% However, most of them don't explore several tasks.
Various approaches~\cite[e.g.,][]{kumar-etal-2020-data, AnabyTavor2020DoNH,mekala-etal-2021-coarse2fine} leverage this capability for \emph{generative data augmentation}. This process usually involves first fine-tuning the GLM on training samples prepended with their target label and then generating synthetic data by prompting the GLM with a given target label.
However, it is not evident that the model parameters learnt during pretraining or fine-tuning should support data generation using such unintuitive formulations with label tokens as prompts: In low data regimes, fine-tuning can be unstable~\cite{devlin-etal-2019-bert} and relies on the pretrained parameters to be reasonably well-suited for the target task~\cite{Phang2018SentenceEO}.
% \tu{I'd get rid of "encoder" here since different model architectures can be used, e.g., encoder-decoder, decoder-only}
Therefore, for target domains that are different from the pretraining domain, such formulations may result in poor quality generation~\cite{feng-etal-2020-genaug}.
% \tu{This could remind reviewers of the target-task language model fine-tuning approach (Howard and Ruder, 2018; Gururangan et al., 2020), which we do not compare against.}

% Lengthier documents are vital for topic classification and are more frequent in real-world scenarios.
% ~\cite{vu-etal-2021-strata} uses natural language inference (NLI) as the auxiliary task and train a GLM for a sentence-level data augmentation.
% % that generates synthetic NLI data, which are usually short.
% ~\cite{mekala-etal-2021-coarse2fine} generates synthetic training data for fine-grained labels, however, they require coarsely-grained annotated dataset as supervision.
% \tu{Consider using something like "sentence-level" data augmentation}

\begin{figure}[t]
    \center
    \includegraphics[width=\linewidth]{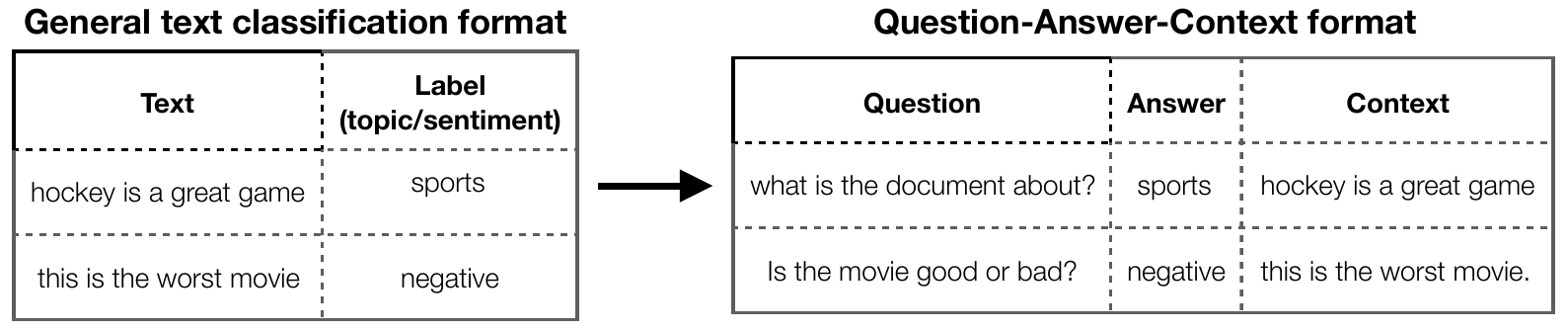}
    % \vspace{-25mm}
    \caption{
    Examples of converting topic classification and sentiment analysis data into question-answer-context format.
    % \tu{I'd remove "topic/sentiment" and shrink the arrow to make the table/text bigger.}
    % \timo{I think we should come up with a different figure here. In my opinion, the figure on the top right of the first page is the thing people look at first, so it should convey the key idea in a really clear, intuitive and easy-to-understand way. The current figure looks a bit boring to me.}
    % \jingbo{it is not very clear in this picture if LOPS is only replacing the ``train'' or ``train + bootstrap''. Maybe I'm too picky.}
    }
    \label{fig:example_qac}
\end{figure}

\begin{figure*}[t]
    \center
    \includegraphics[width=\linewidth]{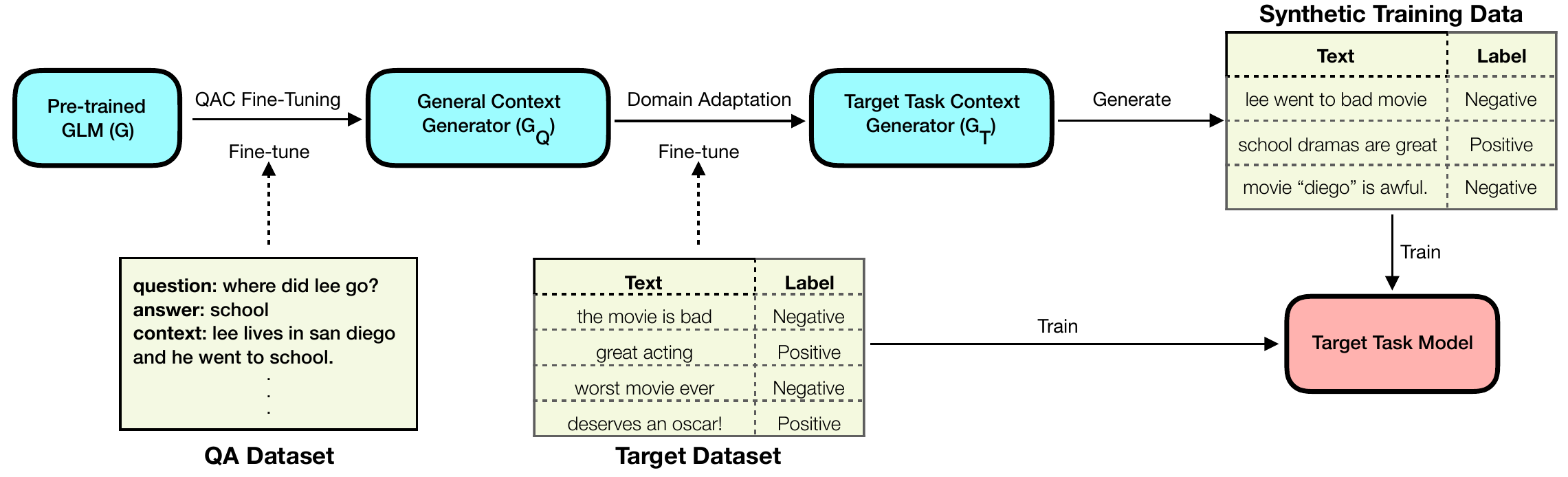}
    % \vspace{-25mm}
    \caption{We propose to use QA datasets for transforming pre-trained generative language models into high-quality target task data generators.
    We view QA datasets in question-answer-context format and fine-tune a pre-trained GLM ($G$) to obtain a general context generator ($G_Q$). Then, we adapt it to the target domain by training it further on few-shot target dataset supervision, resulting in $G_T$. Finally, using $G_T$, we generate synthetic training data for the target task, use it to augment the few-shot target dataset and train the target task model on the augmented data.
    % \timo{Could we maybe use some color to make this diagram a bit easier to understand at first sight?}
    % \jingbo{it is not very clear in this picture if LOPS is only replacing the ``train'' or ``train + bootstrap''. Maybe I'm too picky.}
    }
    \label{fig:overview}
\end{figure*}

% However, generating data for a particular task has been a challenging problem.
% \jingbo{maybe break this paragraph into two. Discussing the challenge a bit more and start a new paragraph for our proposal.}\tu{Why it is challenging given recent successes from prior work. I'd discuss and cite more relevant work that uses generative data augmentation here.}

% \todo{explain the name of method}
To address this challenge, we propose \textbf{\our}, an approach to leverage existing QA datasets for training \emph{\textbf{Con}text generators} to improve generative \textbf{D}ata \textbf{A}ugmentation.
We propose to use a question answering (QA) formulation as a consistent format to prompt GLMs for synthetic data: We use QA datasets for training GLMs to be \emph{context generators} for a given question and answer.
% \tu{QA datasets -> existing QA datasets, to be -> to obtain}
% To address this challenge, we propose to use a question answering (QA) formulation as a consistent format to prompt GLMs for synthetic data: We use QA datasets for intermediate training and finetune GLMs to be \emph{context generators} for a given question and answer~\cite{gardner2019question}.

% To address this challenge, we propose \textbf{\our}, an approach to leverage QA datasets for improved generative \textbf{Con}textual \textbf{D}ata \textbf{A}ugmentation.
% We propose to use a question answering (QA) formulation as a consistent format to prompt GLMs for synthetic data: 
% % We use QA datasets for intermediate training and finetune GLMs to be \emph{context generators} for a given question and answer~\cite{gardner2019question}.
% We use QA datasets for training GLMs to be \emph{context generators} for a given question and answer.
% % To address this challenge, we propose to use a question answering (QA) formulation as a consistent format to prompt GLMs for synthetic data: We use QA datasets for intermediate training and finetune GLMs to be \emph{context generators} for a given question and answer~\cite{gardner2019question}.

As illustrated in Figure~\ref{fig:overview}, our method consists of two steps. The first step is QAC fine-tuning, where we fine-tune a pretrained language model on a QA dataset to obtain a general context generator that is capable of generating contexts for given questions and answers. To this end, we view the QA dataset in question-answer-context format instead of the context-question-answer format used to solve QA tasks~\cite{Radford2018ImprovingLU, Radford2019LanguageMA, 2020t5}.
Then, we adapt the general context generator to the target domain by further training it on available few-shot data, resulting in a target-domain context generator.
Inspired from recent work in converting several NLP tasks into a common format~\cite{McCann2018decaNLP, 2020t5}, we format the target tasks into a question-answer schema.
For example, as shown in Figure~\ref{fig:example_qac}, topic classification and sentiment analysis data can be cast into the question-answer-context format with  
its respective \textit{label} as \textit{answer}, and \textit{text} as \textit{context}.
% Then, we format target sentiment and topic classification datasets into a question-answer schema 
We adapt the context generator to the target task domain by further training on target task few-shot supervision, resulting in target task context generator.
Finally, we generate synthetic training data for the target task by generating contexts for questions and answers pertaining to the respective dataset.
% \tu{context should be in the same single/plural form as the following question \& answer}
Then, we add the generated samples to the few-shot supervision and train a target task model on the augmented data.

We perform extensive experiments on multiple sentiment analysis and topic classification datasets with several abstractive, extractive, and common-sense reasoning QA datasets.
Through rigorous experiments and thorough analysis, we observe that QA datasets that require high-level reasoning abilities such as abstractive and common-sense QA datasets suit the best for generating high-quality data.

Our contributions are summarized as follows:
\begin{itemize}[leftmargin=*,nosep]
    \item We propose to use QA datasets for training generative language models to be context generators for a given question and answer.
    \item We formulate various classification tasks into a QA format and model synthetic training data generation for these tasks as context generation.
    % \tu{The first two contributions can be combined into a single one}
    \item We perform experiments on multiple sentiment analysis and topic classification datasets to demonstrate the effectiveness of our method in zero- and few-shot settings. 
    % \tu{Maybe it's worth mentioning the analysis here, like our analysis sheds light on ...}
    \item We release the code on Github\footnote{\url{https://github.com/dheeraj7596/CONDA}}.
    % \tu{Releasing pre-trained models can be another contribution here}
\end{itemize}
% \timo{I removed the claim about releasing code and datasets, as this would overcomplicate the interal review. We can revisit this later.}
% \noindent\textbf{Reproducibility.} We will release the code on Github\footnote{\url{https://github.com/dheeraj7596/CONDA}}.\tu{I'd get rid of this and mention it as another contribution instead}
% \newpoint{lack of supervision and huge effort on training data}
% \newpoint{Using question-answering format, fine-tune the generative LM}
% \newpoint{Compare it with intermediate-task fine tuning}
% \newpoint{briefly explain the findings}
% \newpoint{write about self-training}
\section{Related Work}
% \timo{Is there a particular reason why related work is the penultimate chapter? Personally, I always prefer to have it directly after the introduction, but that might just be me.}
% \timo{For in-line citations, we should use ``citet'' instead of ``cite'' to have the parentheses placed correctly}
% \timo{In general, I think the related work section might profit from some restructuring and I think there are some works missing. But I don't think we need to address this before the EMNLP deadline}

\paragraph{Data Augmentation}
% \tu{Use \\ paragraph\{Data agumentation:\} to save space. Similar to other subsections.}
% \timo{I removed the first sentence because I don't think we need to explain what data augmentation is used for}
%Data Augmentation has been an active research area to minimize the human annotation effort.
\citet{wei-zou-2019-eda} propose a simple data augmentation method using synonym replacement, random insertion, random swap, and random deletion. 
\citet{sennrich-etal-2016-improving} augment samples by translating them into foreign language and then back to English.
\citet{du-etal-2021-self} compute task-specific query embeddings to retrieve sentences from unlabeled documents from the Internet.
After a rise in pretrained generative language models, the generation capabilities of these models have been explored to generate synthetic data.
\citet{AnabyTavor2020DoNH, kumar-etal-2020-data, schick-schutze-2021-generating, mekala-etal-2021-coarse2fine} generate labeled documents using the GLMs and \cite{yang-etal-2020-generative} do so specifically for common-sense reasoning.
\citet{puri-etal-2020-training} use GLMs to synthesize questions and answers and improve performance on question answering.
\citet{vu-etal-2021-strata} generate data for NLI tasks.
% \timo{Again, I removed the last sentence because I don't understand how that sentence fits into the related work section}
%Our method trains GLMs to be context generators that generates relevant context for a given question and answer.

\paragraph{Few-shot Learning}
% \todo{Add text-to-text papers, formulating as question-answer papers}
Our work is closely related to few-shot learning as we take a few annotated samples as supervision.
The idea of formulating classification as a prompting task is getting increasingly popular.
~\citet{NEURIPS2020_1457c0d6} introduce a new paradigm called in-context learning to infer from large language models using few annotated samples.
% \citet{bansal-etal-2020-self} propose a self-supervised meta-learning approach for few-shot classification. 
\citet{schick-schutze-2021-exploiting} formulate input samples as cloze-style phrases and assign pseudo-labels that are used for training the classifier and \citet{tam-etal-2021-improving} improves their approach further without using any task-specific unlabeled data.
~\cite{McCann2018decaNLP, 2020t5} format several NLP tasks into a question-answer and text-to-text schema.
\citet{lin2021few} train multilingual autoregressive language models to enable few-shot learning in multiple languages.
~\citet{gao-etal-2021-making} propose to generate prompts and convert smaller pretrained language models to few-shot learners. Other work proposes to pre-train prompts by adding soft prompts into the pre-training stage~\cite{gu-etal-2022-ppt,vu-etal-2022-spot,vu-etal-2022-overcoming}.
% \timo{I removed the last sentence because I don't understand how that sentence fits into the related work section}
%We use QA datasets for intermediate training and generate synthetic training data to perform few-shot classification.

\paragraph{Language Model Fine-Tuning}
Pre-trained language models are applied to downstream tasks by fine-tuning them using task-specific objectives~\cite{howard-ruder-2018-universal}.
However, this process requires significant annotated downstream task data~\cite{Yogatama2019LearningAE}.
Many methods have been proposed to address this challenge.
~\citet{gururangan-etal-2020-dont} propose to continue training on unlabeled data from the target task domain. 
~\citet{aghajanyan-etal-2021-muppet} propose pre-finetuning, a large-scale multi-task learning stage between language model pre-training and fine-tuning.
% ~\citet{Gunel2021SupervisedCL} introduce a contrastive learning objective for the fine-tuning.
% ~\citet{Mosbach2021OnTS} propose a strategy stabilize fine-tuning of BERT-based models.
~\citet{Phang2018SentenceEO} introduce intermediate task fine-tuning which involves fine-tuning a language model on an auxiliary task before continuously training on the target task.
~\citet{pruksachatkun-etal-2020-intermediate} observe that the tasks requiring high-level inference and reasoning abilities are the best choice as intermediate tasks.
~\citet{vu-etal-2020-exploring} identify the best auxiliary tasks for high performance on downstream tasks.
% \tu{If space matters, you can copy/write something similar to this sentence from my SPoT paper for the above three papers " Prior
% work shows effective transfer from data-rich source
% tasks~\cite{Phang2018SentenceEO}, those that require complex reasoning and inference~\cite{pruksachatkun-etal-2020-intermediate}, or those that are similar to the target task~\cite{vu-etal-2020-exploring}."}
~\citet{vu-etal-2021-strata} use NLI as auxiliary task to generate synthetic NLI data for intermediate fine-tuning.
% Our method is close to intermediate task fine-tuning, however, we use the fine-tuned generative language model to generate synthetic data instead of training directly for the downstream tasks.
Our method differs from~\cite{Phang2018SentenceEO} in two fronts: (1) we use QA datasets for training context generators instead of answering the question
, and (2) we use the fine-tuned GLM to generate synthetic data instead of training directly for the downstream tasks.
It also differs from ~\cite{vu-etal-2021-strata} in terms of the generated data, where they consider NLI as an auxiliary task and generate synthetic samples in target-domain for the NLI task irrespective of the target task and perform intermediate task fine-tuning.
\our formats target tasks into question-answer format and directly generates samples relevant for target task.
\section{\our: QA Datasets for Generative Data Augmentation}
In this section, we describe the problem statement, and explain our method including QAC fine-tuning, target-domain adaptation, and synthetic training data generation.

\subsection{Problem Formulation}
For a given task $\mathcal{T}$, the input in a few-shot setting contains a very small labeled dataset $\mathcal{L}_\mathcal{T} = \{(\mathcal{D}_1, l_1), (\mathcal{D}_2, l_2), \ldots, (\mathcal{D}_{|\mathcal{L}_\mathcal{T}|}, l_{|\mathcal{L}_\mathcal{T}|})$\} 
% (2) large corpus of unlabeled documents $\mathcal{U}_\mathcal{T} = \{\mathcal{D}_1, \mathcal{D}_2, \ldots, \mathcal{D}_{|\mathcal{U}_\mathcal{T}|}$\}, 
and $m$ target classes $\mathcal{C}$ = \{$\mathcal{C}_1, \mathcal{C}_2, \ldots, \mathcal{C}_m$\}.
% The unlabeled data set $\mathcal{U}_\mathcal{T}$ can be artificially created by removing the ground truth labels of labeled documents.
Our method requires users to provide a question per dataset that is representative of the task to be solved.
% \todo{write about question representing task}
Our aim is to build a model for the task $\mathcal{T}$ that assigns a label $\mathcal{C}_j \in \mathcal{C}$ to each document $\mathcal{D}$. 
% \timo{It said $D_i$ here, I replaced it with a more general $D$ because we do not want to only assign labels to the training data, but to arbitrary documents $D$.}

% \newpoint{Explain the problem, input output}
% \subsection{Intermediate Training on QA Datasets}
\subsection{QAC Fine-tuning}
% \timo{Not every QA dataset has this format, so I slightly reworded this}
We consider question-answering datasets $Q$ containing triplets $(q,a,c)$ of a question $q$, the corresponding answer $a$, and a context $c$ required to derive the correct answer.
Question-answering datasets can roughly be divided into \emph{extractive}~\cite{rajpurkar-etal-2016-squad, trischler-etal-2017-newsqa, joshi-etal-2017-triviaqa, reddy2019coqa} and \emph{abstractive} datasets~\cite{kocisky-etal-2018-narrativeqa, huang-etal-2019-cosmos, xiong2019tweetqa, sap-etal-2019-social}.
% \timo{Would be good to have a few citations here for extractive and abstractive datasets, respectively.}
For extractive QA datasets, the answer can be found as a contiguous span in the context, whereas in abstractive QA datasets, the answer needs to be generated in natural language without being able to rely on the vocabulary of the question or context.
% \tu{I'd merge this sentence with the previous sentence. You can just explain what is extractive/abstractive in the previous sentence}
% \timo{I removed some text here as I don't really think the information is necessary}
%All common-sense QA datasets are abstractive in nature.
%Since we aim to train a context generator, we don't require span information and only consider questions, answers, and contexts for all QA datasets.

We transform the QA dataset $Q$ into training data $D_{QAC}$ for fine-tuning GLM.
To this end, each triplet ($q$, $a$, $c$) is converted into a single text by prepending \textit{``question:''}, \textit{``answer:''} and \textit{``context:''}, respectively, and concatenating $q$, $a$ and $c$ separated by newlines.
% \timo{ Note that the boxes violate the margin rules, this should be fixed.}
For example, a preprocessed training document in $D_{QAC}$ from an extractive QA dataset might look as follows:\\

\noindent\fbox{%
    % \parbox[t]{7.5cm}{
    % \parbox{\columnwidth}{
    \parbox{0.95\linewidth}{
        \textcolor{red}{\texttt{question}}: when did battle of plassey happen?\\
        \textcolor{red}{\texttt{answer}}: 23 june 1757 \\
        \textcolor{red}{\texttt{context}}: the battle of plassey was a decisive victory of the british east india company over the nawab of bengal and his french allies on 23 june 1757.
    }
}
\\

We fine-tune a pretrained GLM $G$ on $D_{QAC}$ to obtain a \emph{general context generator} $G_Q$ using a language modeling objective to maximize the log-likelihood of the ($q$, $a$, $c$) triplet. 
The general context generator $G_Q$ is capable of generating contexts for given questions
and answers.
% \timo{I don't think the mathematical part here is necessary as the NLL formulation should be known to the reader. Also, I don't think the equation is correct. Why would $(P(q_i, a_i, c_i; \Theta))$ be the same as $P(c_i|q_i, a_i; \Theta)$?}
%Mathematically, for a given question $q_i$, answer $a_i$, and context $c_i$, we learn parameters $\Theta$ by maximizing $\mathcal{L}$ where:
%\begin{align*}
%    \mathcal{L} =  \sum_ilog(P(q_i, a_i, c_i; \Theta)) \\
%    = \sum_ilogP(c_i|q_i, a_i; \Theta)
%\end{align*}
%One can view our formulation as asking the question $q_i$ and answer $a_i$ to play the role of prompt and the context
%$c_i$ to be the continuation, thus facilitating conditional context generation.
% We call this step QAC fine-tuning.

% This language modeling loss facilitates conditional generation of context given question and answer, which we discuss in detail in next section.
% In our experiments, we use GPT2-medium~\cite{Radford2019LanguageMA} as $G$.

\subsection{Domain Adaptation and Synthetic Training Data Generation}
% \newpoint{Context generator and using questions and answers}

We adopt $G_Q$ to the target domain by fine-tuning it further on available few-shot data.
To preserve its context generating ability, we perform QAC fine-tuning instead of regular language model fine-tuning.
This is enabled by transforming the few-shot supervision into our question-answer-context format.
First, we manually design one question per dataset that is representative of the task and the dataset. 
Furthermore, following ~\citet{schick-schutze-2021-exploiting}, we define a verbalizer as a mapping $v$: $\mathcal{C} \rightarrow \mathcal{V}$ that maps each label in $\mathcal{C}$ to a word from $G_Q$’s vocabulary $\mathcal{V}$.
Finally, for every document $\mathcal{D}_i$ and its respective label $l_i$ in our few-shot data, we consider the verbalizer mapping of the label, $v(l_i)$, as answer and the text $\mathcal{D}_i$ as context.
For example, a \textit{negative} review \textit{``I hate this movie''} from the IMDb dataset \citep{maas-EtAl:2011:ACL-HLT2011} is transformed as follows:\\

\noindent\fbox{%
    % \parbox[t]{7.5cm}{
    % \parbox{\columnwidth}{
    \parbox{0.95\linewidth}{
        \textcolor{red}{\texttt{question}}: is the movie good or bad?\\
        \textcolor{red}{\texttt{answer}}: bad \\
        \textcolor{red}{\texttt{context}}: i hate this movie.
    }
}
\\

We fine-tune $G_Q$ on the converted few-shot data to obtain a target task context generator $G_\mathcal{T}$.

\paragraph{Synthetic Training Data Generation} Recall that our method requires a question $q$ for every dataset that is representative of the task to be solved.
To obtain synthetic training data, for every distinct label $\mathcal{C}_j$, we create a question-answer prompt with $q$ as question, $v(\mathcal{C}_j)$ as answer and let $G_\mathcal{T}$ generate the context $c_{gen}$.
% \timo{I'm pretty sure that the following equation is wrong - obtaining the argmax would be intractable, which is why we would practically use some approximation like greedy decoding or beam search. But then again, we don't do that either as we sample from our distribution. I would just remove this equation alltogether.}
%Specifically, given question $q$ and answer $v(\mathcal{C}_j)$, we conditionally generate $c_{gen}$ using $P(c_{gen}|q, a)$ probabilities from $G_\mathcal{T}$:
%\begin{align*}
%    c_{gen} = \arg\max_c P(c | q, a)
%\end{align*}
The generated context $c_{gen}$ and label $\mathcal{C}_j$ are considered as a synthetic training sample. 
We repeat this process multiple times to generate $n$ samples that we collect in a synthetic training dataset denoted by $\mathcal{D}_{gen}$.

As a final step, we train the target task model on the combination of $\mathcal{D}_{gen}$ and our original few-shot dataset $\mathcal{L}_\mathcal{T}$.
We use this trained target-task model for inference. 
\section{Experiments}
In this section, we evaluate our method against several data augmentation and few-shot methods on sentiment analysis and text classification tasks. 

\begin{table}[t]
    \center
    \scalebox{0.73}{
    \begin{tabular}{l l r l}
        \toprule
        {\textbf{Dataset}} & {\textbf{Type}} & {\textbf{\# Samples}} & {\textbf{Training Context}} \\
        \midrule
        \textbf{SQuAD} & Extractive & 87,600 & Wikipedia \\
        \textbf{NewsQA} & Extractive & 76,560 & News\\
        \textbf{TweetQA} & Abstractive & 10,692 & News Tweets\\
        \textbf{SocialIQA} & Commonsense & 33,410 & Crowdsourcing \\
        \textbf{CosmosQA} & Commonsense & 21,448 & Blogs \\
        \bottomrule
    \end{tabular}
    }
    \caption{Relevant statistics of the QA dataset used in our experiments.}
    \label{tbl:qa_datastats}
\end{table}

\subsection{QA Datasets}
We consider several extractive, abstractive, and common-sense QA datasets.
Common-sense QA datasets are also abstractive datasets that require common-sense reasoning to answer the questions.
The QA dataset statistics are provided in Table~\ref{tbl:qa_datastats}.
The details of these datasets are as follows:
% \tu{Can rewrite the whole paragraph as: "We consider several .... datasets, including:" and get rid of this sentence}
\begin{itemize}[leftmargin=*,nosep]
    \item \textbf{SQuAD}~\cite{rajpurkar-etal-2016-squad, rajpurkar-etal-2018-know} is a collection of questions and answers based on Wikipedia articles.
    \item \textbf{NewsQA}~\cite{trischler-etal-2017-newsqa} is a challenging QA dataset in the News domain where crowdworkers were shown a news article’s headline and summary, and asked to formulate a question about the article without accessing its content.
    \item \textbf{TweetQA}~\cite{xiong2019tweetqa} is a QA dataset made from a collection of tweets sampled from two major news websites (CNN and NBC). 
    \item \textbf{SocialIQA}~\cite{sap-etal-2019-social} is a QA dataset that tests social common-sense intelligence. The data is made of common phrases from stories and books. 
    \item \textbf{CosmosQA}~\cite{huang-etal-2019-cosmos}  is a commonsense-based reading comprehension task formulated as multiple-choice questions. Answering questions requires reasoning not only based on the exact text spans in the context, but also abstractive commonsense reasoning.
    % \tu{I'd get rid of all "is a QA dataset" in the above descriptions since we mentioned that before and the dataset names speak for themselves.}
\end{itemize}

\begin{table}[t]
    \center
    % \caption{Dataset statistics. \jingbo{Make it two-column and add all your final label names into the table?}}
    \small
     \scalebox{0.65}{
    \begin{tabular}{c c c}
        \toprule
            {\textbf{Dataset}} & {\textbf{Question}} & {\textbf{Verbalized Labels}}  \\
        \midrule
        \multicolumn{3}{c}{\textbf{Sentiment}}\\
        \midrule
        IMDb & \textit{is this movie good or bad?} &  \begin{tabular}{@{}c@{}}  good, bad \end{tabular}\\
        \cmidrule{1-3}
        Yelp & \textit{how is the service?} & \begin{tabular}{@{}c@{}} awful, bad, fine,\\  good, excellent \end{tabular}\\
        \cmidrule{1-3}
        SST-2 & \textit{is this sentence positive or negative?} & \begin{tabular}{@{}c@{}}  positive, negative \end{tabular}\\
        \midrule
        \multicolumn{3}{c}{\textbf{Topic}}\\
        \midrule
        Yahoo & \textit{what is this document about?} &  \begin{tabular}{@{}c@{}}  sports, society, science, health, \\ politics, education, computer, \\  business, entertainment, relationship \end{tabular}\\
        \cmidrule{1-3}
        NYT & \textit{what is this document about?} & \begin{tabular}{@{}c@{}} arts, business, \\  politics, sports \end{tabular}\\
        \cmidrule{1-3}
        AGNews & \textit{what is this document about?} & \begin{tabular}{@{}c@{}}  sports, business, \\  technology, politics \end{tabular}\\
        \bottomrule
    \end{tabular}
    }
    \caption{Questions and Verbalized labels of the target task datasets considered in our experiments.
    % \tu{This table is hard to read, put "sentiment" as the second row spanning all the columns (similar for "topic"). That way you can save a large space to make the text bigger}
    }
    \label{tbl:target_datastats}
\end{table}

\subsection{Target Task Datasets}
We evaluate our method on six English text classification datasets. In particular, we consider the three sentiment analysis datasets: IMDb reviews~\cite{maas-EtAl:2011:ACL-HLT2011}, Yelp\footnote{\url{https://www.yelp.com/dataset/}}, and SST-2~\cite{socher-etal-2013-recursive}, as well as three topic classification datasets: Yahoo~\cite{zhang2015characterlevel}, The New York Times\footnote{\url{http://developer.nytimes.com/}} (NYT), and AGNews~\cite{zhang2015characterlevel}.
% \timo{We should add a citation here for Yelp reviews and NYT.}
The dataset-representative questions, and their respective verbalized labels of target task datasets are mentioned in Table~\ref{tbl:target_datastats}.
We follow and adapt ~\citet{McCann2018decaNLP} for questions in sentiment analysis datasets. The question for topic classification is intuitive and straightforward.
% The statistics of sentiment analysis datasets is mentioned in Table~\ref{tbl:sent_datastats}. 
More details about the datasets can be found in Appendix~\ref{app:target_task_datas}.

% \tu{I'd list all of these dataset within a single paragraph to save space. They are all common datasets so we don't need to introduce them much.}
% \begin{table}[t]
%     \center
%     \caption{Topic Classification Dataset statistics.}
%     \label{tbl:topic_datastats}
%     \vspace{-3mm}
%     % \small
%     \begin{tabular}{c c c c}
%         \toprule
%         {\textbf{Dataset}} & {\textbf{\# Test Examples}} & {\textbf{\# Labels}} \\
%         \midrule
%         \textbf{Yahoo} & 60000 & 10 \\
%         \textbf{NYT} & 10582 & 5 \\
%         \textbf{AGNews} & 114000 & 4 \\
%         \bottomrule
%     \end{tabular}
% \end{table}

% \subsection{Topic Classification Datasets}
% The statistics of sentiment analysis datasets is mentioned in Table~\ref{tbl:topic_datastats}. The details of the datasets are as follows:
% \begin{itemize}[leftmargin=*,nosep]
%     \item \textbf{Yahoo:}~\cite{zhang2015characterlevel} is a topic classification dataset with question and answer pairs. Using these pairs, the task is to predict their corresponding topic.
%     \item \textbf{The New York Times (NYT):} : contains news articles written and published by The New York Times that are classified into 5 wide genres.
%     \item \textbf{AGNews:}~\cite{zhang2015characterlevel} is a topic categorization dataset in news domain from AG’s corpus.
% \end{itemize}

\subsection{Compared Methods}
We compare with a wide range of data augmentation and intermediate-task fine-tuning (ITFT) methods described below:
\begin{itemize}[leftmargin=*,nosep]
    \item \textbf{BERT-FT} trains the \texttt{BERT-base-uncased} classifier~\cite{devlin-etal-2019-bert} on the few-shot supervision.
    \item \textbf{ITFT-$X$}~\cite{Phang2018SentenceEO} first trains a model on dataset $X$ and fine-tunes it further on the target task. We compare with ITFT-MNLI and ITFT-SQuAD fine-tuned intermediately on MNLI~\cite{N18-1101} and SQuAD datasets respectively.
    \item \textbf{BackTranslation}~\cite{sennrich-etal-2016-improving} augments samples by translating them into a non-English language and translating them back to English. We translate them to French, Spanish, and Portuguese thereby augmenting three synthetic samples for every sample.
    \item \textbf{PEGASUS}~\cite{zhang2019pegasus} is a state-of-the-art paraphrasing model. We paraphrase the input text and consider it as a synthetic sample and augment it to the training set.
    \item \textbf{EDA}~\cite{wei-zou-2019-eda} generates synthetic samples by synonym replacement, random insertion, random swap, and random deletion and augment them to the training set.
    \item \textbf{LAMBADA}~\cite{AnabyTavor2020DoNH} fine-tunes a GLM on few-shot supervision prepended with their target labels and then generates synthetic data by prompting the GLM with a given target label.
\end{itemize}

We denote our method as \our, which includes QAC fine-tuning, domain adaptation, synthetic samples generation, and training the target task classifier.
\our-$X$ represents that the QAC fine-tuning of GLM is performed on QA dataset $X$.
We also compare with $\our\backslash QA$ where we perform no QAC fine-tuning and directly fine-tune the GLM on target dataset. 
% \tu{GLM -> the GLM} \tu{Can rewrite this as: For ablation studies, we include X where we...}
% We also compare with an ablation $\our\backslash QA$ where we perform no QAC fine-tuning on QA datasets and directly fine-tune GLM on target dataset. 
% \timo{I think we shouldn't use omissions in scientific writing (such as ``don't'' instead of ``do not''.}

\subsection{Experiment Settings} 
We consider two low-data regimes: few-shot and zero-shot.
We consider 8 annotated samples per label in the few-shot setting.
In the zero-shot setting, we skip the domain adaptation step and use $G_Q$ directly for synthetic training data generation and train the target task model only on the generated synthetic training data.
We use GPT2-Medium~\cite{Radford2019LanguageMA} as our GLM and fine-tune it for 3 epochs in QAC-fine-tuning and domain adaptation steps.
While generating synthetic training samples, we use top-$k$ sampling with $k = 20$, a maximum length of 200 tokens, and generate $n = 450$ synthetic samples per label.
We use \texttt{BERT-base-uncased}~\cite{devlin-etal-2019-bert} as target task classifier.
We feed \texttt{[CLS]} representation into the classification head and train all the parameters on the downstream target tasks.
% \tu{Maybe can rewrite the above two sentences as: For fine-tuning BERT on a downstream classification task, we ...}
Following ~\cite{devlin-etal-2019-bert}, we fix the number of epochs of target task BERT classifier training to 4 unless mentioned otherwise.
We perform 3 random restarts and report the mean and standard deviation.\footnote{For each restart, we resample the few-shot training set.}
We use the Transformers library~\cite{wolf-etal-2020-transformers} and NVIDIA RTX A6000 GPUs for our experiments. 

To enable a fair comparison, we generate the same number of samples per label as \our (i.e., 450) for all data augmentation baselines. 
We use \texttt{BERT-base-uncased} as the target task classifier for all baselines.
$\our\backslash QA$ for zero-shot setting implies a pre-trained GPT2.
While training the target task classifier, since the number of training samples for baselines like BERT-FT, ITFT are different than data augmentation baselines and our method \our, we set the number of epochs for all baselines such that the number of update steps remain the same for a fair comparison.

\begin{table*}[t]
    \center
    \scalebox{0.67}{
    % \small
    \begin{tabular}{l cccccc cccccc}
        \toprule
                         & \multicolumn{6}{c}{\textbf{Sentiment}} & \multicolumn{6}{c}{\textbf{Topic}} \\
        % \cmidrule{2-13}
        \cmidrule(lr){2-7} \cmidrule(lr){8-13}
                         & \multicolumn{2}{c}{\textbf{IMDb}} & \multicolumn{2}{c}{\textbf{Yelp}} & \multicolumn{2}{c}{\textbf{SST-2}} &  \multicolumn{2}{c}{\textbf{NYT}} & \multicolumn{2}{c}{\textbf{Yahoo}} & \multicolumn{2}{c}{\textbf{AGNews}}
                         \\
        \textbf{Method} & Mi-F$_1$ & Ma-F$_1$ & Mi-F$_1$ & Ma-F$_1$ & Mi-F$_1$ & Ma-F$_1$ & Mi-F$_1$ & Ma-F$_1$ & Mi-F$_1$ & Ma-F$_1$ & Mi-F$_1$ & Ma-F$_1$ \\
        \midrule
        BERT-FT & $69.1_{4.9}$ & $69.1_{4.9}$ & $39.8_{2.3}$ & $38.9_{3.4}$ & $62.0_{4.7}$ & $61.8_{4.8}$ & $94.4_{1.1}$ & $88.1_{1.6}$ & $55.4_{2.1}$ & $55.2_{1.6}$ & $78.4_{1.8}$ & $78.3_{1.8}$ \\
        ITFT-MNLI & \underline{$73.9_{4.6}$} & \underline{$73.5_{4.8}$} & $40.4_{2.6}$ & $40.0_{2.9}$ & \underline{$66.5_{6.9}$} & \underline{$65.5_{6.9}$} & $90.1_{1.2}$ & $80.8_{1.1}$ & $38.7_{5.8}$ & $37.6_{5.2}$ & $71.1_{1.1}$ & $70.6_{1.1}$ \\
        ITFT-SQuAD & $65.5_{4.4}$ & $64.6_{4.3}$ & $38.4_{2.5}$ & $36.8_{2.5}$ & $61.9_{2.2}$ & $61.5_{2.2}$ & $93.0_{0.5}$ & $85.3_{1.3}$ & $45.3_{3.4}$ & $45.0_{3.0}$ & $72.0_{1.9}$ & $71.4_{2.1}$\\
        BackTranslation & $68.0_{4.6}$ & $67.1_{5.2}$ & \underline{$41.6_{2.4}$} & \underline{$40.3_{3.0}$} & $60.6_{4.5}$ & $60.0_{4.9}$ & $95.4_{0.6}$ & $90.0_{1.3}$ & $57.4_{1.4}$ & $57.1_{1.2}$ & $80.0_{2.2}$ & $79.8_{2.3}$ \\
        PEGASUS & $66.8_{5.0}$ & $65.9_{5.6}$ & $35.9_{3.7}$ & $34.4_{3.1}$ & $61.1_{5.3}$ & $60.9_{5.3}$ & $93.2_{0.5}$ & $87.2_{0.6}$ & $58.1_{1.9}$ & $57.2_{1.7}$ & \underline{$81.1_{1.9}$} & \underline{$80.9_{2.1}$}\\
        EDA & $63.6_{1.4}$ & $62.1_{1.6}$ & $39.1_{1.9}$ & $37.9_{2.0}$ & $57.4_{4.0}$ & $52.9_{7.4}$ & \underline{$95.8_{0.8}$} & \underline{$90.9_{1.7}$} & $56.1_{1.7}$ & $55.8_{1.8}$ & $80.0_{3.0}$ & $79.8_{3.0}$\\
        LAMBADA & $50.3_{0.7}$ & $42.3_{5.4}$ & $20.8_{1.06}$ & $11.1_{6.3}$ & $49.6_{1.3}$ & $45.8_{3.3}$ & $60.3_{19.7}$ & $45.9_{17.4}$ & $25.7_{4.7}$ & $22.6_{3.2}$ & $49.3_{9.9}$ & $46.9_{10.3}$ \\
        $\our\backslash QA$ & $72.2_{6.9}$ & $71.3_{8.0}$ & $36.8_{0.6}$ & $23.9_{1.7}$ & $50.6_{0.5}$ & $35.1_{0.5}$ & $93.5_{0.8}$ & $85.7_{1.2}$ & \underline{$58.5_{0.3}$} & \underline{$57.3_{0.4}$} & $79.4_{1.6}$ & $78.8_{1.8}$ \\
        \midrule
        \our-SQuAD & $53.9_{1.9}$ & $45.9_{6.2}$ & $37.9_{0.7}$ & $31.1_{3.2}$ & $51.5_{1.6}$ & $39.8_{7.6}$ & $93.2_{0.8}$ & $86.0_{1.7}$ & $56.9_{0.5}$ & $55.4_{0.5}$ & $\mathbf{81.6_{0.8}}$ & $\mathbf{81.3_{0.9}}$ \\
        \our-NewsQA & $57.9_{3.7}$ & $55.5_{5.4}$ & $36.4_{1.1}$ & $31.6_{2.0}$ & $56.0_{6.2}$ & $50.5_{10.3}$ & $91.5_{0.3}$ & $81.1_{0.6}$ & $58.3_{0.7}$ & $57.2_{0.8}$ & $80.0_{3.3}$ & $79.6_{3.6}$ \\
        \our-TweetQA & $\mathbf{75.1_{2.3}}$ & $\mathbf{74.5_{2.5}}$ & $\mathbf{42.9_{1.1}}$ & $\mathbf{42.0_{1.8}}$ & $\mathbf{67.7_{4.8}}$ & $\mathbf{67.5_{4.9}}$ & $94.1_{0.6}$ & $86.6_{1.3}$ & $\mathbf{59.4_{0.4}}$ & $\mathbf{58.1_{0.3}}$ & $\mathbf{83.0_{0.9}}$ & $\mathbf{82.9_{0.9}}$  \\
        \our-SocialIQA & $\mathbf{79.5_{1.9}}$ & $\mathbf{79.5_{1.9}}$ & $39.4_{1.5}$ & $32.2_{2.8}$ & $\mathbf{75.4_{1.4}}$ & $\mathbf{75.2_{1.6}}$ & $93.2_{3.4}$ & $85.8_{1.3}$ & $\mathbf{61.9_{0.5}}$ & $\mathbf{61.1_{0.6}}$ & $\mathbf{81.9_{0.2}}$ & $\mathbf{81.7_{0.2}}$\\
        \our-CosmosQA & $\mathbf{77.0_{3.2}}$ & $\mathbf{76.4_{3.7}}$ & $\mathbf{42.3_{0.1}}$ & $37.5_{1.0}$ & $\mathbf{67.4_{0.6}}$ & $\mathbf{66.9_{1.2}}$ & $94.3_{0.4}$ & $87.7_{1.1}$ & $\mathbf{63.8_{0.6}}$ & $\mathbf{63.3_{0.4}}$ & $\mathbf{82.8_{0.8}}$ & $\mathbf{82.5_{0.8}}$ \\
        \bottomrule
    \end{tabular}
    }
    \caption{Few-Shot Evaluation Results. Micro- and Macro-F1 are used as evaluation metrics. All experiments are repeated with three random seeds. Mean and standard deviation (in the subscript) are reported. The best baseline for each dataset is underlined and all results of \our that outperform the best baseline are in bold. 
    % \tu{I'd report macro F1 only. Using both metrics adds nothing but makes it more complicated and hard to read the table.}
    }
    % \timo{I think using both micro and macro F1 makes this unnecessarily complicated (as they both point in the exact same direction). Can we use just one of them?}\tu{Agreed! I'd remove one to make the table more readable.}
    \label{tbl:f1_results}
\end{table*}

\begin{table*}[t]
    \center
    \scalebox{0.67}{
    % \small
    \begin{tabular}{l cccccc cccccc}
        \toprule
                         & \multicolumn{6}{c}{\textbf{Sentiment}} & \multicolumn{6}{c}{\textbf{Topic}} \\
        % \cmidrule{2-13}
        \cmidrule(lr){2-7} \cmidrule(lr){8-13}
                         & \multicolumn{2}{c}{\textbf{IMDb}} & \multicolumn{2}{c}{\textbf{Yelp}} & \multicolumn{2}{c}{\textbf{SST-2}} &  \multicolumn{2}{c}{\textbf{NYT}} & \multicolumn{2}{c}{\textbf{Yahoo}} & \multicolumn{2}{c}{\textbf{AGNews}}
                         \\
        \textbf{Method} & Mi-F$_1$ & Ma-F$_1$ & Mi-F$_1$ & Ma-F$_1$ & Mi-F$_1$ & Ma-F$_1$ & Mi-F$_1$ & Ma-F$_1$ & Mi-F$_1$ & Ma-F$_1$ & Mi-F$_1$ & Ma-F$_1$ \\
        \midrule
        $\our\backslash QA$ & $70.9_{3.7}$ & $70.0_{3.8}$ & $32.4_{3.7}$ & $19.8_{2.0}$ & $52.9_{3.6}$ & $41.8_{9.9}$ & $90.7_{0.9}$ & $80.0_{1.8}$ & $57.9_{0.3}$ & $57.1_{0.7}$ & $77.0_{1.5}$ & $76.2_{1.6}$\\
        \our-SQuAD & $53.3_{2.4}$ & $42.7_{7.1}$ & $30.2_{4.5}$ & $21.4_{4.6}$ & $52.4_{2.6}$ & $47.3_{5.9}$ & $85.7_{3.8}$ & $74.3_{3.1}$ & $56.5_{1.5}$ & $54.9_{1.7}$ & $79.3_{0.1}$ & $78.9_{0.2}$\\
        \our-NewsQA & $53.4_{2.6}$ & $47.4_{10.0}$ & $32.8_{1.0}$ & $23.1_{3.6}$ & $51.6_{1.7}$ & $46.2_{3.6}$ & $89.8_{0.2}$ & $77.1_{1.1}$ & $55.9_{1.1}$ & $54.6_{0.8}$ & $76.8_{3.0}$ & $75.7_{3.5}$\\
        \our-TweetQA & $72.4_{5.0}$ & $70.6_{6.1}$ & $\mathbf{38.0_{2.4}}$ & $\mathbf{37.6_{2.3}}$ & $61.2_{3.9}$ & $56.5_{6.4}$ & $90.3_{1.9}$ & $78.5_{4.7}$ & $54.0_{0.4}$ & $52.4_{0.2}$ & $76.4_{1.7}$ & $76.2_{2.0}$\\
        \our-SocialIQA & $\mathbf{78.3_{4.3}}$ & $\mathbf{77.6_{5.1}}$ & $36.9_{2.2}$ & $31.7_{1.1}$ & $\mathbf{76.2_{1.2}}$ & $\mathbf{76.1_{1.2}}$ & $87.0_{3.0}$ & $77.6_{2.9}$ & $56.1_{1.6}$ & $55.2_{2.1}$ & $79.6_{1.9}$ & $79.4_{2.2}$\\
        \our-CosmosQA & $75.9_{2.3}$ & $75.5_{2.8}$ & $37.4_{1.5}$ & $35.1_{1.7}$ & $66.2_{6.4}$ & $66.0_{6.5}$ & $\mathbf{92.8_{0.3}}$ & $\mathbf{84.5_{1.3}}$ & $\mathbf{63.4_{0.5}}$ & $\mathbf{62.9_{0.3}}$ & $\mathbf{81.8_{1.6}}$ & $\mathbf{81.5_{1.4}}$\\
        \bottomrule
    \end{tabular}
    }
    \caption{Zero-Shot Evaluation Results. Mean and standard deviation (in the subscript) are reported.
    % \tu{I'd report macro F1 only. Using both metrics adds nothing but makes it more complicated and hard to read the table.}
    }
    \label{tbl:f1_zero_shot}
\end{table*}

\subsection{Results and Discussion}
Results for few- and zero-shot settings are shown in Table~\ref{tbl:f1_results} and Table~\ref{tbl:f1_zero_shot}, respectively, using Micro- and Macro-F1 as evaluation metrics. We discuss the effectiveness of our method below.

\smallsection{\our vs Baselines} In the few-shot setting, \our with abstractive and common-sense based datasets outperforms all baselines for most of the datasets, beating the best baseline in five out of six cases.
\our performs better than BERT-FT on all datasets, achieving up to $14\%$ improvement on SST-2.
Although ITFT performs better than vanilla fine-tuning, \our demonstrates better performance than ITFT on all datasets.
For example, \our-TweetQA shows $11\%$ improvement over ITFT-SQuAD on AG-News.
\our demonstrates higher performance than data-augmentation baselines on all datasets except NYT.
The comparison between \our and LAMBADA shows that our QA formulation prompt is more intuitive and informative than just the target label.
%NYT is relatively easier dataset that results in near supervised performance (94\% micro f1) with BERT-FT on just 8 samples per label.
% \timo{I removed a sentence about NYT because I don't think it's that relevant here}
We attribute the superior performance of \our to the context-generating ability acquired during QAC fine-tuning that is efficiently leveraged by generating synthetic samples, which are added to the training set.

\smallsection{Abstractive vs Extractive QA Datasets}
We observe that the performance of \our with abstractive QA datasets is significantly better than \our with extractive QA datasets in both few-shot and zero-shot settings.
For example, \our-TweetQA has an improvement of more than $20\%$ over \our-SQuAD on IMDb in few-shot setting.
We surmise that this is because of the intrinsic nature of extractive QA datasets (i.e., the answer always being present in the context as a contiguous span).
We observe that GLMs fine-tuned on an extractive QA dataset retain the ability to generate contexts that encompass the answer. 
Note that, while generating synthetic training samples, the answer in the prompt is its respective topic.
For example, out of $500$ generated samples by \our-SQuAD for Yelp dataset, $213$ samples contain at least one occurrence of its corresponding verbalized label whereas it is only $73$ for \our-CosmosQA.
Thus, many synthetic samples generated contain their corresponding label in text.
Therefore, a classifier trained on synthetic samples that have their corresponding labels in the text, easily overfits on the label tokens and does not generalize well to unseen test data.

\smallsection{Comparison with $\our\backslash$QA}
\our with abstractive QA datasets perform better than $\our\backslash QA$ in both few-shot and zero-shot settings, attaining improvements up to $40\%$ and $35\%$ respectively
% $40\%$ in few-shot and $42\%$ in zero-shot setting 
in macro-F1 on SST-2. 
This demonstrates that the context generating abilities are learnt and reinforced during the QAC fine-tuning on QA datasets which is efficiently utilized by generating synthetic samples.

\smallsection{Zero-shot Performance} The zero-shot performance of \our follows a similar trend as few-shot performance: abstractive and common-sense reasoning QA datasets lead to better performance than extractive datasets and no QAC fine-tuning. 
%This demonstrates that our method results in high quality context generators that are able to generate artificial training data that results in high performance without any human annotated data.  

% \smallsection{Validation vs No Validation Set} 
% We seldom observe significant improvement upon introducing the validation set. 
% This is because a small validation set which is of same size as few-shot supervision is not large enough to tune the hyperparameters.
\begin{table}[t]
    \center
    \scalebox{0.53}{
    % \small
    \begin{tabular}{c c ccc ccc}
        \toprule
                         & & \multicolumn{3}{c}{\textbf{Sentiment}} & \multicolumn{3}{c}{\textbf{Topic}} \\
        % \cmidrule{2-13}
        \cmidrule(lr){3-5} \cmidrule(lr){6-8}
                         \textbf{QA Dataset} & \textbf{Setting} & {\textbf{IMDb}} & {\textbf{Yelp}} & {\textbf{SST-2}} &  {\textbf{NYT}} & {\textbf{Yahoo}} & {\textbf{AGNews}}
                         \\
        % \textbf{QA Dataset} & \textbf{Setting} & Ma-F$_1$ & Ma-F$_1$ & Ma-F$_1$ & Ma-F$_1$ & Ma-F$_1$ & Ma-F$_1$ \\
        \midrule
        \multirow{3}{*}{SQuAD} & \our & $45.9_{6.2}$ & $31.1_{3.2}$ & $39.8_{7.6}$ & $86.0_{1.7}$ & $55.4_{0.5}$ & $81.3_{0.9}$ \\
                               & - DA & $51.3_{12.7}$ & $28.2_{0.5}$ & $33.4_{0.1}$ & $87.1_{1.0}$ & $55.0_{1.7}$ & $82.5_{0.6}$\\
                               & - Few Shot & $49.4_{9.5}$ & $25.9_{3.4}$ & $43.7_{4.0}$ & $75.0_{4.0}$ & $47.4_{0.4}$ & $77.9_{3.0}$\\
                            %   & Zero shot & $42.7_{7.1}$ & $21.4_{4.6}$ & $47.3_{5.9}$ & $74.3_{3.1}$ & $54.9_{1.7}$ & $78.9_{0.2}$\\
        \midrule
        \multirow{3}{*}{NewsQA} & \our & $55.5_{5.4}$ & $31.6_{2.0}$ & $50.5_{10.3}$ & $81.1_{0.6}$ & $57.2_{0.8}$ & $79.6_{3.6}$ \\
                                & - DA & $60.9_{9.6}$ & $32.0_{3.6}$ & $46.2_{8.5}$ & $79.8_{0.4}$ & $56.4_{1.1}$ & $79.2_{3.5}$\\
                                & - Few Shot & $50.9_{4.8}$ & $23.4_{1.9}$ & $46.0_{7.0}$ & $77.2_{2.1}$ & $54.3_{0.7}$ & $76.0_{4.1}$\\
                                % & Zero shot & 
        \midrule
        \multirow{3}{*}{TweetQA} & \our & $74.5_{2.5}$ & $42.0_{1.8}$ & $67.5_{4.9}$ & $86.6_{1.3}$ & $58.1_{0.3}$ & $82.9_{0.9}$  \\
                              & - DA & $80.5_{3.5}$ & $42.1_{0.2}$ & $63.2_{7.5}$ & $85.3_{2.1}$ & $57.1_{1.1}$ & $81.1_{1.6}$\\
                              & - Few Shot & $74.0_{2.7}$ & $40.6_{0.7}$ & $59.3_{12.4}$ & $77.3_{4.8}$ & $53.8_{0.3}$ & $77.4_{1.5}$\\
        \midrule
        \multirow{3}{*}{SocialIQA} & \our & $79.5_{1.9}$ & $32.2_{2.8}$ & $75.2_{1.6}$ & $85.8_{1.3}$ & $61.1_{0.6}$ & $81.7_{0.2}$\\
                              & - DA & $81.0_{1.9}$ & $35.3_{0.8}$ & $76.1_{0.6}$ & $87.6_{1.3}$ & $60.7_{0.8}$ & $82.4_{1.2}$\\
                              & - Few Shot & $77.7_{4.0}$ & $31.4_{3.0}$ & $75.2_{1.0}$ & $76.5_{1.7}$ & $55.8_{1.2}$ & $78.2_{1.4}$\\
        \midrule
        \multirow{3}{*}{CosmosQA} & \our & $76.4_{3.7}$ & $37.5_{1.0}$ & $66.9_{1.2}$ & $87.7_{1.1}$ & $63.3_{0.4}$ & $82.5_{0.8}$ \\
                              & - DA & $76.3_{2.4}$ & $36.8_{2.3} $ & $51.8_{9.3}$ & $ 87.1_{1.0}$ & $62.9_{0.3}$ & $82.6_{0.6}$\\
                              & - Few Shot & $74.9_{1.0}$ & $35.8_{1.4}$ & $64.4_{9.1}$ & $82.9_{0.7}$ & $60.1_{0.2}$ & $80.3_{1.4}$\\
        \bottomrule
    \end{tabular}
    }
    \caption{Ablation Study. Macro-F1 is used as evaluation metric.}
    \label{tbl:f1_ablation}
\end{table}

\subsection{Ablation Study}
To understand the impact of domain adaptation and few-shot samples, we compare \our with two ablated versions in Table~\ref{tbl:f1_ablation}: (1) \textit{\our-DA} represents our method without domain adaptation (i.e., generating synthetic data using $G_Q$ and training the classifier on combined few-shot supervision and synthetic data generated by $G_Q$), (2) \textit{\our-Few Shot} represents the classifier trained only on the samples generated by $G_T$.
We also present the results of our complete pipeline for reference.
\our performs better than \our-Few shot in most cases, demonstrating the importance of including few-shot samples in the training set for the classifier.
The comparison between \our and \our-DA suggests that fine-tuning the language model further on the target dataset helps in some scenarios but does not always improve performance.
This is in line with previous research findings~\cite{du-etal-2021-self, vu-etal-2021-strata, Pryzant2022AutomaticRI}.
We conjecture that domain adaptation is important when the structure of the target task dataset is very different from the QA dataset. For example, domain adaptation helps most of the QA datasets on SST-2 dataset because the text in SST-2 is a single sentence, whereas most of the QA datasets have paragraphs as context.
Moreover, it also depends on the number of samples the language model is fine-tuned on during domain adaptation.
We observe that the higher the number of samples, the more positive their impact. 
For example, the number of few-shot samples is the highest in Yahoo compared to other datasets and domain adaptation positively contributes to the performance on Yahoo for all QA datasets.

\subsection{Larger Generative Language Models}
\begin{table}[t]
    \center
    \scalebox{0.53}{
    % \small
    \begin{tabular}{c ccc ccc}
        \toprule
                         & \multicolumn{3}{c}{\textbf{Sentiment}} & \multicolumn{3}{c}{\textbf{Topic}} \\
        % \cmidrule{2-13}
        \cmidrule(lr){2-4} \cmidrule(lr){5-7}
                       \textbf{Method} & {\textbf{IMDb}} & {\textbf{Yelp}} & {\textbf{SST-2}} &  {\textbf{NYT}} & {\textbf{Yahoo}} & {\textbf{AGNews}} \\
        % \textbf{Method} & Mi-F$_1$ & Ma-F$_1$ & Mi-F$_1$ & Ma-F$_1$ & Mi-F$_1$ & Ma-F$_1$ & Mi-F$_1$ & Ma-F$_1$ & Mi-F$_1$ & Ma-F$_1$ & Mi-F$_1$ & Ma-F$_1$ \\
        \midrule
        $\our\backslash QA$-L & $79.7_{2.1}$ & $43.2_{4.4}$ & $67.6_{8.2}$ & $84.8_{1.3}$ & $60.3_{0.8}$ & $77.7_{2.2}$\\
        \midrule
        \our-L-SQuAD & $70.0_{15.2}$ & $38.9_{1.5}$ & $64.3_{7.2}$ & $84.3_{2.9}$ &$60.3_{1.3}$ & $\mathbf{81.1_{1.7}}$ \\
        \our-L-NewsQA & $72.4_{9.4}$ & $38.3_{0.3}$ & $58.1_{6.2}$ & $\mathbf{85.5_{2.4}}$ & $\mathbf{61.4_{1.1}}$ & $\mathbf{82.2_{1.2}}$ \\
        \our-L-TweetQA & $76.4_{5.7}$ & $\mathbf{45.0_{1.3}}$ & $\mathbf{74.6_{2.3}}$ & $84.4_{0.1}$ & $\mathbf{61.6_{0.1}}$ & $\mathbf{79.7_{3.1}}$ \\
        \our-L-SocialIQA & $\mathbf{81.6_{2.4}}$ & $\mathbf{43.9_{3.4}}$ & $\mathbf{77.5_{1.3}}$ & $\mathbf{89.0_{0.4}}$ & $\mathbf{62.0_{0.5}}$ & $\mathbf{80.9_{2.2}}$ \\
        \our-L-CosmosQA & $\mathbf{83.4_{0.6}}$ & $\mathbf{43.2_{2.2}}$ & $\mathbf{77.2_{1.7}}$ & $\mathbf{86.5_{2.4}}$ & $\mathbf{61.0_{0.6}}$ & $\mathbf{79.5_{3.9}}$ \\
        \bottomrule
    \end{tabular}
    }
    \caption{Few-Shot Evaluation Results with GPT2-Large as GLM (\textit{-L} denotes GPT2-Large). Macro-F1 is used as evaluation metric. All results of \our-L that perform better than $\our\backslash QA$-L are in bold.}
    \label{tbl:f1_results_large}
\end{table}

Experimental results with GPT2-Large as the GLM are shown in Table~\ref{tbl:f1_results_large}. 
We observe that the relative performance trend remains the same as GPT2-Medium i.e. \our with abstractive datasets performs better than \our with extractive datasets and $\our\backslash QA$-L.
This indicates that QAC fine-tuning improves the performance of generative data augmentation with larger GLMs as well.

\subsection{Performance vs No. of Generated Samples}

\begin{figure}[t]
    \subfigure[AGNews]{
        \includegraphics[width=0.47\linewidth]{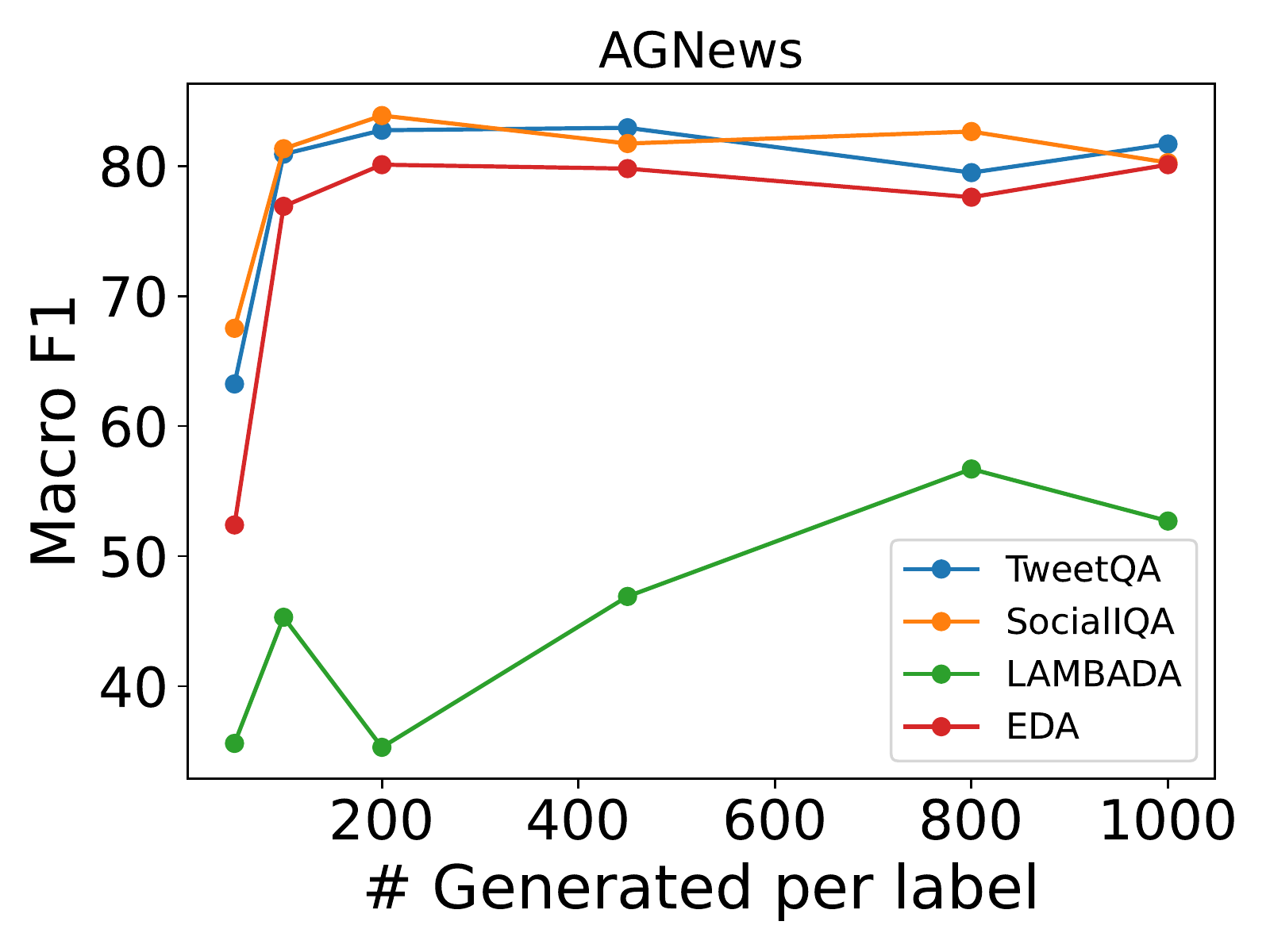}
    }
    \subfigure[IMDb]{
        \includegraphics[width=0.47\linewidth]{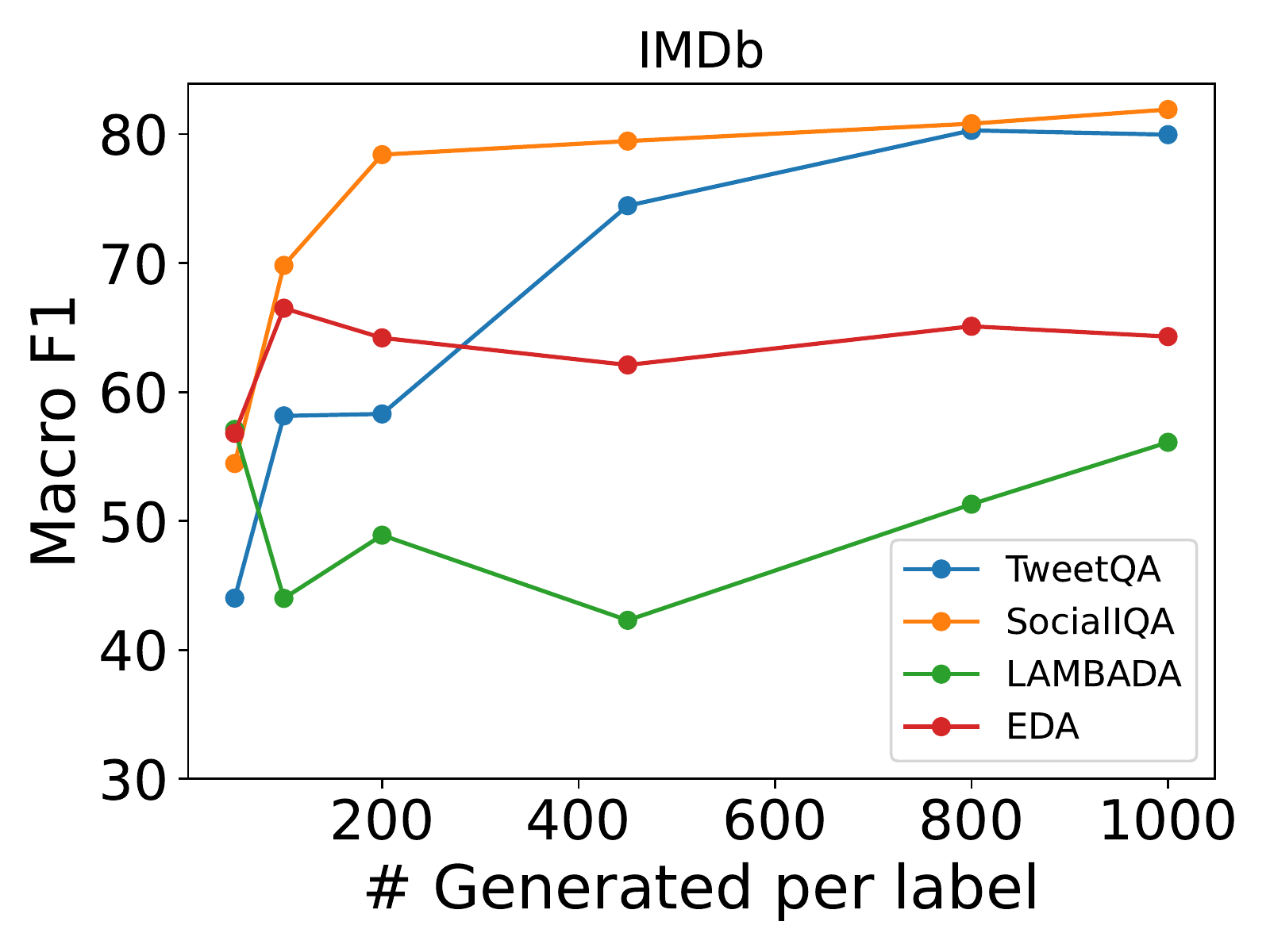}
    }
    \vspace{-3mm}
    \caption{Macro-F$_1$ scores of \our-TweetQA and \our-SocialIQA w.r.t. number of generated samples per class. We fix the few-shot supervision size to 8 samples per label. Each experiment is repeated with three different seeds and the mean performance is plotted.}
    \label{figure:perf_gen}
    \vspace{-5mm}
\end{figure}

% \newpoint{Plot the performance vs number of generated samples and explain}
We fix the few-shot supervision size to 8 samples per label and vary the number of generated samples per label and plot the performance of \our-TweetQA, \our-SocialIQA, and baselines such as LAMBADA and EDA on AGNews and IMDb datasets, shown in Figure~\ref{figure:perf_gen}. 
We repeat each setting with three different seeds and plot the mean performance.
We observe that the performance increases and it plateaus after a while.
This shows that synthetic training data can give a substantial boost to the few-shot training data, minimizing the human effort in manual annotations; however, it cannot replace the original training data completely as it requires more human annotated data to improve beyond some limit.

\subsection{Performance vs Few-shot supervision Size}
\begin{figure}[t]
    \subfigure[SST-2]{
        \includegraphics[width=0.47\linewidth]{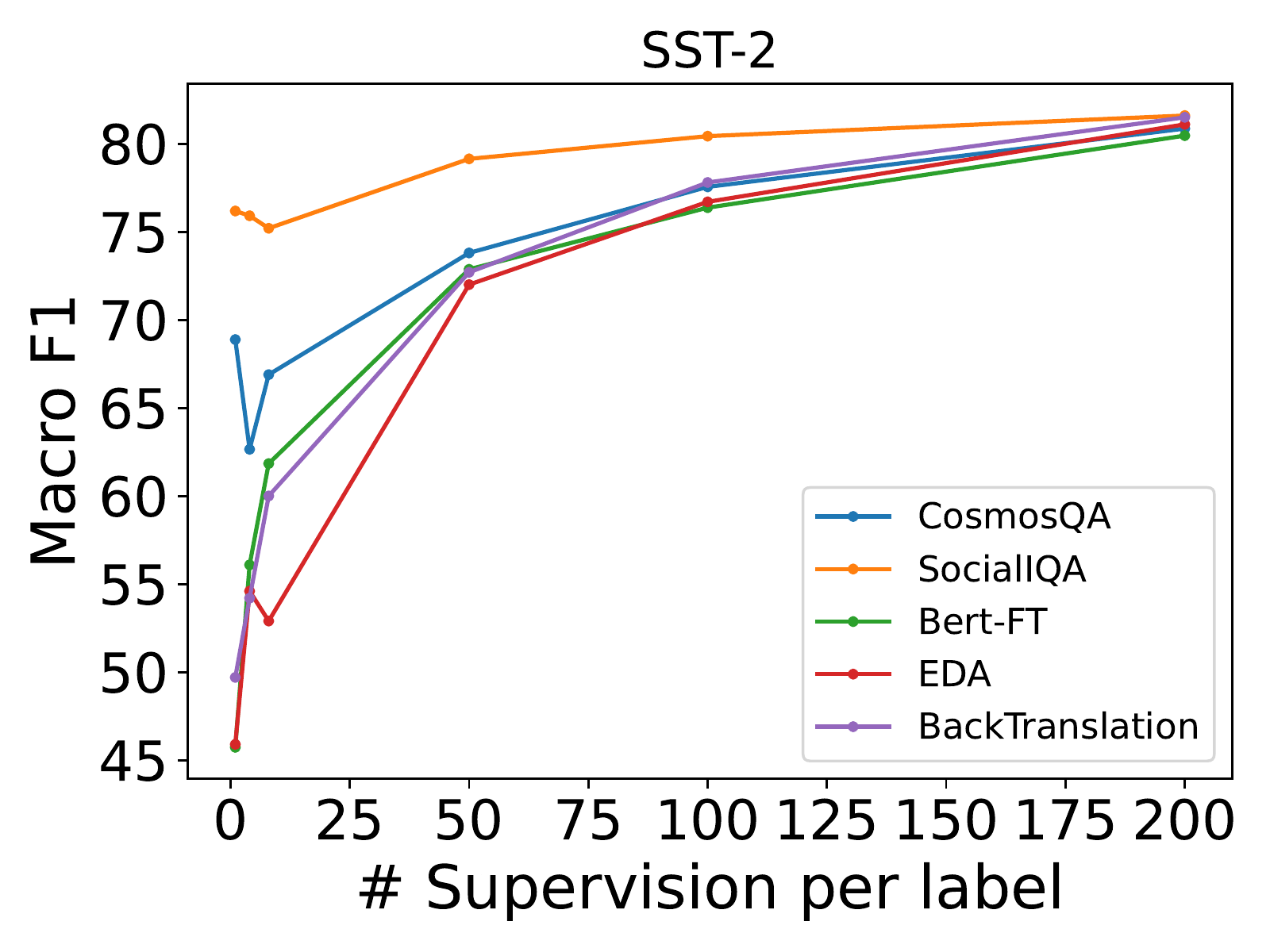}
    }
    \subfigure[Yahoo]{
        \includegraphics[width=0.47\linewidth]{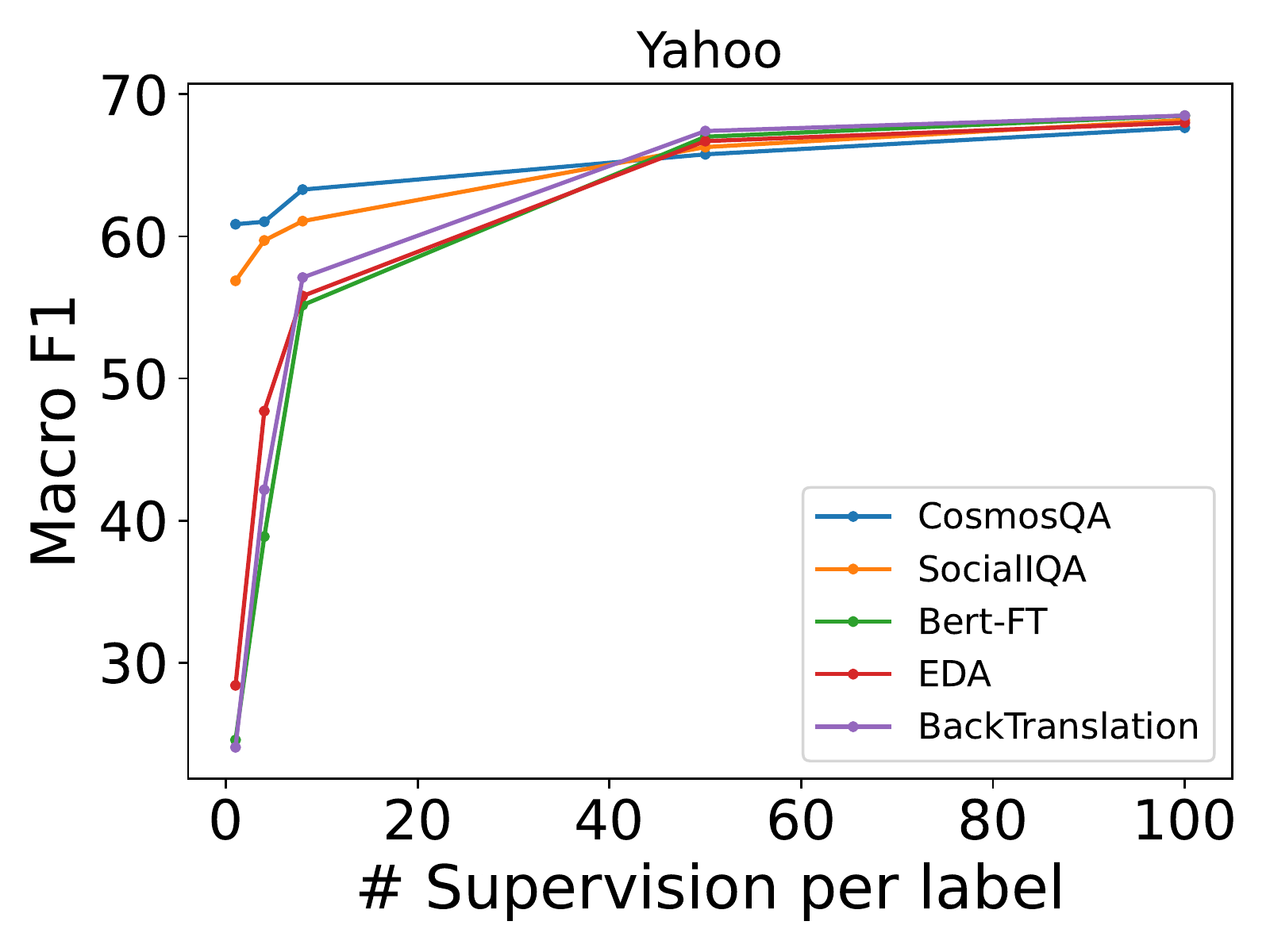}
    }
    \vspace{-3mm}
    \caption{Macro-F$_1$ scores of \our-CosmosQA and \our-SocialIQA w.r.t. number of few-shot annotated samples per class. Each experiment is repeated with three different seeds and mean performance is plotted.}
    \label{figure:perf_sup}
    \vspace{-5mm}
\end{figure}

We fix the number of generated samples to 450 per label and vary the number of annotated samples and plot the performance of \our-CosmosQA and \our-SocialIQA on SST-2 and Yahoo datasets in Figure~\ref{figure:perf_sup}.
We also plot the performance of baselines such as BERT-FT, EDA, BackTranslation for comparison.
We repeat each experiment with three random seeds and plot the mean performance.
We observe that the performance of \our increases with the size of supervision and the improvement over baselines in the low-data regime is substantial. 
For example, with only 4 annotated samples per label in Yahoo dataset, the macro F1 of \our-CosmosQA outperforms BERT-FT by $22\%$ and EDA by $15\%$.
However, we also observe that the performance gap between \our and baselines decreases with increase in supervision size and gets stagnated after a while.
As the size of supervision increases, the supervision by itself is sufficient for high performance, thus reducing the performance boost due to synthetic training data.
% As also observed by ~\cite{pruksachatkun-etal-2020-intermediate}, we conjecture that with the increase in size of supervision, the QAC fine-tuned generative language model loses its context-generating ability when fine-tuned further on many samples during domain adaptation.
% \todo{rewrite this}

\begin{table}[t]
    \center
    \scalebox{0.71}{
    % \small
    \begin{tabular}{c cc cc}
        \toprule
                        %  & \multicolumn{3}{c}{\textbf{Sentiment}} & \multicolumn{3}{c}{\textbf{Topic}} \\
        % \cmidrule{2-13}
        % \cmidrule(lr){2-4} \cmidrule(lr){5-7}
                       \textbf{Method} & {\textbf{IMDb}} & {\textbf{SST-2}} &  {\textbf{Yahoo}} & {\textbf{AGNews}} \\
        % \textbf{Method} & Mi-F$_1$ & Ma-F$_1$ & Mi-F$_1$ & Ma-F$_1$ & Mi-F$_1$ & Ma-F$_1$ & Mi-F$_1$ & Ma-F$_1$ & Mi-F$_1$ & Ma-F$_1$ & Mi-F$_1$ & Ma-F$_1$ \\
        \midrule
        LMPT & $64.8_{3.6}$ & $57.5_{2.1}$ & $49.9_{1.2}$ & $79.2_{1.7}$\\
        \midrule
        \our-TweetQA & $\mathbf{74.5_{2.5}}$ & $\mathbf{67.5_{4.9}}$ & $\mathbf{58.1_{0.3}}$ & $\mathbf{82.9_{0.9}}$ \\
        \our-SocialIQA & $\mathbf{79.5_{1.9}}$ & $\mathbf{75.2_{1.6}}$ & $\mathbf{61.1_{0.6}}$ & $\mathbf{81.7_{0.2}}$ \\
        \our-CosmosQA & $\mathbf{76.4_{3.7}}$ & $\mathbf{66.9_{1.2}}$ & $\mathbf{63.3_{0.4}}$ & $\mathbf{82.5_{0.8}}$ \\
        \bottomrule
    \end{tabular}
    }
    \caption{Few-Shot Evaluation comparison between language model pre-training on unlabeled data (LMPT) and \our. Macro-F1 is used as evaluation metric. All results of \our that perform better than LMPT are in bold.}
    \label{tbl:f1_results_lmpt}
\end{table}

\subsection{Self-Training}
\begin{table}[t]
    \center
    \scalebox{0.8}{
    \small
    \begin{tabular}{c c c c c}
        \toprule
        %                  & & \multicolumn{3}{c}{\textbf{Sentiment}} & \multicolumn{3}{c}{\textbf{Topic}} \\
        % % \cmidrule{2-13}
        % \cmidrule(lr){3-5} \cmidrule(lr){6-8}
                         \textbf{QA Dataset} & \textbf{Setting} & {\textbf{SST-2}} &  {\textbf{NYT}} & {\textbf{AGNews}}
                         \\
        % \textbf{QA Dataset} & \textbf{Setting} & Ma-F$_1$ & Ma-F$_1$ & Ma-F$_1$ \\
        \midrule
        % \multirow{5}{*}{SQuAD} & \our &  \\
        %                       & \our + ST &\\
        %                       \cmidrule{2-5}
        %                       & \our-L & $64.3_{7.2}$ \\
        %                       & \our-L + ST &  \\
        % \midrule
        % \multirow{5}{*}{NewsQA} & \our &  \\
        %                       & \our + ST &\\
        %                       \cmidrule{2-5}
        %                       & \our-L & $58.1_{6.2}$  \\
        %                       & \our-L + ST &  \\
        % \midrule
        \multirow{5}{*}{TweetQA} & \our & $67.5_{4.9}$ & $86.6_{1.3}$ & $82.9_{0.9}$ \\
                               & \our + ST & $\mathbf{69.2_{1.3}}$ & $\mathbf{88.2_{1.0}}$ & $82.4_{1.7}$\\
                               \cmidrule{2-5}
                               & \our-L & $74.6_{2.3}$ & $84.4_{0.1}$ & $79.7_{3.1}$\\
                               & \our-L + ST & $\mathbf{76.9_{1.1}}$ & $\mathbf{87.4_{2.4}}$ & $\mathbf{80.9_{3.4}}$ \\
        \midrule
        \multirow{5}{*}{SocialQA} & \our & $75.2_{1.6}$ & $85.8_{1.3}$ & $81.7_{0.2}$ \\
                               & \our + ST & $\mathbf{79.8_{0.8}}$ & $\mathbf{90.3_{1.9}}$ & $\mathbf{83.9_{1.5}}$\\
                               \cmidrule{2-5}
                               & \our-L & $77.5_{1.3}$ & $89.0_{0.4}$ & $80.9_{2.2}$\\
                               & \our-L + ST & $\mathbf{78.6_{0.6}}$ & $\mathbf{92.1_{0.8}}$ & $\mathbf{81.1_{1.8}}$\\
        \midrule
        \multirow{5}{*}{CosmosQA} & \our & $66.9_{1.2}$ & $87.7_{1.1}$ & $82.5_{0.8}$ \\
                               & \our + ST & $\mathbf{71.6_{6.9}}$ & $87.2_{2.3}$ & $\mathbf{83.6_{0.6}}$\\
                               \cmidrule{2-5}
                               & \our-L & $77.2_{1.7}$ & $86.5_{2.4}$ & $79.5_{3.9}$\\
                               & \our-L + ST & $\mathbf{79.2_{1.3}}$ & $\mathbf{87.2_{4.0}}$ & $\mathbf{80.7_{3.6}}$\\
        \bottomrule
    \end{tabular}
    }
    \caption{Self-Training experiment results with Macro-F1 as evaluation metric. \emph{+ ST} denotes with self-training. Self-training improves the performance of both \our and \our-L significantly. All results where self-training improved the performance are in bold.}
    \label{tbl:f1_st}
\end{table}

We perform an experiment to demonstrate that the performance can be further improved through self-training when in-domain unlabeled samples are provided. 
In-domain unlabeled samples are often easily available in real-world scenarios.
Self-training is a commonly-used approach to bootstrap the classifier on unlabeled samples~\cite{mekala2020contextualized, mekala2020meta, vu-etal-2021-strata}.
Following ~\citet{vu-etal-2021-strata}, we obtain pseudo-labels by predicting on unlabeled samples using the trained classifier and train the classifier further on the available labeled and pseudo-labeled data. 
We consider the training set without ground truth labels as unlabeled data and experiment on SST-2, NYT, and AGNews datasets.
We repeat this process for 3 iterations without any filtering of pseudo-labels.
From the results in Table~\ref{tbl:f1_st}, we can observe a significant performance improvement up to 4 points with self-training.
It is noteworthy that this improvement is consistent for both GPT2-Medium and Large models respectively.
% \tu{This section lacks of training details for self-training, can just cite STraTA if you follow a similar training procedure/use STraTA code.}

\subsection{Synthetic Data Adds Value}
Unsupervised language model pre-training(LMPT) on target-task unlabeled data can improve performance~\cite{gururangan-etal-2020-dont}.
We consider training set without ground truth labels as unlabeled data for LMPT and present a comparison in few-shot setting in Table~\ref{tbl:f1_results_lmpt}.
We observe \our performs better than LMPT demonstrating the quality and importance of generated synthetic data.

\subsection{Case study: Evaluating Context Generator}
We hypothesize that our method results in high-quality context generators that are capable of generating context for a given question and answer.
To validate this hypothesis in in-domain and out-of-domain settings, we perform two experiments on QA task.

\smallsection{In-domain Analysis}
In this experiment, we validate whether the context generator is capable of generating context for question, answer pairs belonging to the same domain as QA dataset used for QAC fine-tuning.
We consider SQuAD dataset and partition it into training set with 1000 (question, answer, context) triplets, dev set of size 1700 with only (question, answer) pairs and a test set of size 6570.
First, we consider GPT2-Medium as GLM and perform QAC fine-tuning on the training set. Then, we generate contexts for the dev set and augment the (question, answer, generated context) triplets to the training set.
Finally, we train a \texttt{BERT-base-uncased} QA model on the augmented data. 
We compare it with the BERT model trained only on the original training set.
We report F1 scores on test set in Table~\ref{tbl:case_study}. We observe a boost of $4\%$ using our synthetic training data, validating our hypothesis in the in-domain setting.

\smallsection{Out-of-domain Analysis}
In this experiment, we validate our hypothesis in the out-of-domain setting i.e. the domain of target dataset is different than the QA dataset used for QAC fine-tuning.
We follow our proposed pipeline and consider SQuAD as the QA dataset for QAC fine-tuning and NewsQA as the target dataset.
We partition NewsQA dataset into 1000 (question, answer, context) triplets for domain adaptation, 17000 (question, answer) pairs for context generation, and test on 10000 samples.
% We follow our proposed pipeline and consider the SQuAD dataset as the training set and NewsQA as the target dataset. 
We fine-tune GPT2-medium on SQuAD to obtain general context generator and adapt to the NewsQA domain by training it further on 1000 question, answer, context triplets from NewsQA.
Using the target task context generator, we generate contexts for 17000 question, answer pairs, augment it to the training set, and train \texttt{BERT-base-uncased} QA model on the augmented data. 
From F1 scores reported in Table~\ref{tbl:case_study}, we can observe more than $10\%$ improvement in the performance, demonstrating the efficiency of our method in out-of-domain setting.

\begin{table}[t]
    \center
    % \caption{Dataset statistics. \jingbo{Make it two-column and add all your final label names into the table?}}
    \vspace{-3mm}
    \caption{Case Study: We evaluate our context generators in in-domain and out-of-domain settings. In both cases, we observe substantial improvement in the performance demonstrating the effectiveness of our method.}
    \label{tbl:case_study}
    \small
     \scalebox{1}{
    \begin{tabular}{c c c}
        \toprule
            {\textbf{Setting}} & {\textbf{Model}} & {\textbf{F$_1$ score}}  \\
        \midrule
        \multirow{2}{*}{In-domain} & BERT & $32.11$\\
                                   & \our & $36.74$\\
        \midrule
        \multirow{2}{*}{Out-of-domain} & BERT & $14.96$\\
                               & \our & $25.31$\\
        \bottomrule
    \end{tabular}
    }
\end{table}

\section{Conclusion}

In this paper, we propose to train generative language models to be context generators for a given question and answer.
To facilitate this, we use question answer as a format and utilize QA datasets for training generative language models into context generators.
We view sentiment and topic classification tasks in question-answer form and generate contexts using our fine-tuned generative language models.
These generated contexts are used as synthetic training data to augment existing few-shot data for training a classifier.
Extensive experiments on multiple sentiment and topic classification datasets demonstrate strong performance of our method in few-shot and zero-shot settings.

%\section{Acknowledgements}
%We thank Timo Schick for his contribution and valuable discussions.
%Our work is sponsored in part by National Science Foundation Convergence Accelerator under award OIA-2040727 as well as generous gifts from Google, Adobe, and Teradata. Any opinions, findings, and conclusions or recommendations expressed herein are those of the authors and should not be interpreted as necessarily representing the views, either expressed or implied, of the U.S. Government. The U.S. Government is authorized to reproduce and distribute reprints for government purposes not withstanding any copyright annotation hereon.

% \clearpage\newpage
\section{Limitations}
One limitation of our approach is the synthetic training data generated can boost the performance up to an extent and beyond that it requires more annotated samples.
So, the generated synthetic training data cannot replace the training data altogether but could minimize the annotation effort significantly.
Moreover, some tasks such as NER are challenging to cast into question-answering format, which hinders generating synthetic data using our method.
% \timo{I would completely remove the limitations section to save some space, as it doesn't really mention anything that isn't already mentioned elsewhere in the paper.}
\section{Acknowledgements}

We thank anonymous reviewers and program chairs for their valuable and insightful feedback. 
The research was sponsored in part by National Science Foundation Convergence Accelerator under award OIA-2040727 as well as generous gifts from Google, Adobe, and Teradata.
Any opinions, findings, and conclusions or recommendations expressed herein are those of the authors and should not be interpreted as necessarily representing the views, either expressed or implied, of the U.S. Government. 
The U.S. Government is authorized to reproduce and distribute reprints for government purposes not withstanding any copyright annotation hereon.
% \clearpage\newpage
% \section{Ethical Consideration}

\typeout{}
\bibliography{anthology,custom}

\begin{thebibliography}{51}
\expandafter\ifx\csname natexlab\endcsname\relax\def\natexlab#1{#1}\fi

\bibitem[{Aghajanyan et~al.(2021)Aghajanyan, Gupta, Shrivastava, Chen,
  Zettlemoyer, and Gupta}]{aghajanyan-etal-2021-muppet}
Armen Aghajanyan, Anchit Gupta, Akshat Shrivastava, Xilun Chen, Luke
  Zettlemoyer, and Sonal Gupta. 2021.
\newblock \href {https://doi.org/10.18653/v1/2021.emnlp-main.468} {Muppet:
  Massive multi-task representations with pre-finetuning}.
\newblock In \emph{Proceedings of the 2021 Conference on Empirical Methods in
  Natural Language Processing}, pages 5799--5811, Online and Punta Cana,
  Dominican Republic. Association for Computational Linguistics.

\bibitem[{Anaby-Tavor et~al.(2020)Anaby-Tavor, Carmeli, Goldbraich, Kantor,
  Kour, Shlomov, Tepper, and Zwerdling}]{AnabyTavor2020DoNH}
Ateret Anaby-Tavor, Boaz Carmeli, Esther Goldbraich, Amir Kantor, George Kour,
  Segev Shlomov, N.~Tepper, and Naama Zwerdling. 2020.
\newblock Do not have enough data? deep learning to the rescue!
\newblock In \emph{AAAI}.

\bibitem[{Brown et~al.(2020)Brown, Mann, Ryder, Subbiah, Kaplan, Dhariwal,
  Neelakantan, Shyam, Sastry, Askell, Agarwal, Herbert-Voss, Krueger, Henighan,
  Child, Ramesh, Ziegler, Wu, Winter, Hesse, Chen, Sigler, Litwin, Gray, Chess,
  Clark, Berner, McCandlish, Radford, Sutskever, and
  Amodei}]{NEURIPS2020_1457c0d6}
Tom Brown, Benjamin Mann, Nick Ryder, Melanie Subbiah, Jared~D Kaplan, Prafulla
  Dhariwal, Arvind Neelakantan, Pranav Shyam, Girish Sastry, Amanda Askell,
  Sandhini Agarwal, Ariel Herbert-Voss, Gretchen Krueger, Tom Henighan, Rewon
  Child, Aditya Ramesh, Daniel Ziegler, Jeffrey Wu, Clemens Winter, Chris
  Hesse, Mark Chen, Eric Sigler, Mateusz Litwin, Scott Gray, Benjamin Chess,
  Jack Clark, Christopher Berner, Sam McCandlish, Alec Radford, Ilya Sutskever,
  and Dario Amodei. 2020.
\newblock \href
  {https://proceedings.neurips.cc/paper/2020/file/1457c0d6bfcb4967418bfb8ac142f64a-Paper.pdf}
  {Language models are few-shot learners}.
\newblock In \emph{Advances in Neural Information Processing Systems},
  volume~33, pages 1877--1901. Curran Associates, Inc.

\bibitem[{Devlin et~al.(2019)Devlin, Chang, Lee, and
  Toutanova}]{devlin-etal-2019-bert}
Jacob Devlin, Ming-Wei Chang, Kenton Lee, and Kristina Toutanova. 2019.
\newblock \href {https://doi.org/10.18653/v1/N19-1423} {{BERT}: Pre-training of
  deep bidirectional transformers for language understanding}.
\newblock In \emph{Proceedings of the 2019 Conference of the North {A}merican
  Chapter of the Association for Computational Linguistics: Human Language
  Technologies, Volume 1 (Long and Short Papers)}, pages 4171--4186,
  Minneapolis, Minnesota. Association for Computational Linguistics.

\bibitem[{Du et~al.(2021)Du, Grave, Gunel, Chaudhary, Celebi, Auli, Stoyanov,
  and Conneau}]{du-etal-2021-self}
Jingfei Du, Edouard Grave, Beliz Gunel, Vishrav Chaudhary, Onur Celebi, Michael
  Auli, Veselin Stoyanov, and Alexis Conneau. 2021.
\newblock \href {https://doi.org/10.18653/v1/2021.naacl-main.426}
  {Self-training improves pre-training for natural language understanding}.
\newblock In \emph{Proceedings of the 2021 Conference of the North American
  Chapter of the Association for Computational Linguistics: Human Language
  Technologies}, pages 5408--5418, Online. Association for Computational
  Linguistics.

\bibitem[{Feng et~al.(2020)Feng, Gangal, Kang, Mitamura, and
  Hovy}]{feng-etal-2020-genaug}
Steven~Y. Feng, Varun Gangal, Dongyeop Kang, Teruko Mitamura, and Eduard Hovy.
  2020.
\newblock \href {https://doi.org/10.18653/v1/2020.deelio-1.4} {{G}en{A}ug: Data
  augmentation for finetuning text generators}.
\newblock In \emph{Proceedings of Deep Learning Inside Out (DeeLIO): The First
  Workshop on Knowledge Extraction and Integration for Deep Learning
  Architectures}, pages 29--42, Online. Association for Computational
  Linguistics.

\bibitem[{Gao et~al.(2021)Gao, Fisch, and Chen}]{gao-etal-2021-making}
Tianyu Gao, Adam Fisch, and Danqi Chen. 2021.
\newblock \href {https://doi.org/10.18653/v1/2021.acl-long.295} {Making
  pre-trained language models better few-shot learners}.
\newblock In \emph{Proceedings of the 59th Annual Meeting of the Association
  for Computational Linguistics and the 11th International Joint Conference on
  Natural Language Processing (Volume 1: Long Papers)}, pages 3816--3830,
  Online. Association for Computational Linguistics.

\bibitem[{Gu et~al.(2022)Gu, Han, Liu, and Huang}]{gu-etal-2022-ppt}
Yuxian Gu, Xu~Han, Zhiyuan Liu, and Minlie Huang. 2022.
\newblock \href {https://doi.org/10.18653/v1/2022.acl-long.576} {{PPT}:
  Pre-trained prompt tuning for few-shot learning}.
\newblock In \emph{Proceedings of the 60th Annual Meeting of the Association
  for Computational Linguistics (Volume 1: Long Papers)}, pages 8410--8423,
  Dublin, Ireland. Association for Computational Linguistics.

\bibitem[{Gururangan et~al.(2020)Gururangan, Marasovi{\'c}, Swayamdipta, Lo,
  Beltagy, Downey, and Smith}]{gururangan-etal-2020-dont}
Suchin Gururangan, Ana Marasovi{\'c}, Swabha Swayamdipta, Kyle Lo, Iz~Beltagy,
  Doug Downey, and Noah~A. Smith. 2020.
\newblock \href {https://doi.org/10.18653/v1/2020.acl-main.740} {Don{'}t stop
  pretraining: Adapt language models to domains and tasks}.
\newblock In \emph{Proceedings of the 58th Annual Meeting of the Association
  for Computational Linguistics}, pages 8342--8360, Online. Association for
  Computational Linguistics.

\bibitem[{Howard and Ruder(2018)}]{howard-ruder-2018-universal}
Jeremy Howard and Sebastian Ruder. 2018.
\newblock \href {https://doi.org/10.18653/v1/P18-1031} {Universal language
  model fine-tuning for text classification}.
\newblock In \emph{Proceedings of the 56th Annual Meeting of the Association
  for Computational Linguistics (Volume 1: Long Papers)}, pages 328--339,
  Melbourne, Australia. Association for Computational Linguistics.

\bibitem[{Huang et~al.(2019)Huang, Le~Bras, Bhagavatula, and
  Choi}]{huang-etal-2019-cosmos}
Lifu Huang, Ronan Le~Bras, Chandra Bhagavatula, and Yejin Choi. 2019.
\newblock \href {https://doi.org/10.18653/v1/D19-1243} {Cosmos {QA}: Machine
  reading comprehension with contextual commonsense reasoning}.
\newblock In \emph{Proceedings of the 2019 Conference on Empirical Methods in
  Natural Language Processing and the 9th International Joint Conference on
  Natural Language Processing (EMNLP-IJCNLP)}, pages 2391--2401, Hong Kong,
  China. Association for Computational Linguistics.

\bibitem[{Joshi et~al.(2017)Joshi, Choi, Weld, and
  Zettlemoyer}]{joshi-etal-2017-triviaqa}
Mandar Joshi, Eunsol Choi, Daniel Weld, and Luke Zettlemoyer. 2017.
\newblock \href {https://doi.org/10.18653/v1/P17-1147} {{T}rivia{QA}: A large
  scale distantly supervised challenge dataset for reading comprehension}.
\newblock In \emph{Proceedings of the 55th Annual Meeting of the Association
  for Computational Linguistics (Volume 1: Long Papers)}, pages 1601--1611,
  Vancouver, Canada. Association for Computational Linguistics.

\bibitem[{Ko{\v{c}}isk{\'y} et~al.(2018)Ko{\v{c}}isk{\'y}, Schwarz, Blunsom,
  Dyer, Hermann, Melis, and Grefenstette}]{kocisky-etal-2018-narrativeqa}
Tom{\'a}{\v{s}} Ko{\v{c}}isk{\'y}, Jonathan Schwarz, Phil Blunsom, Chris Dyer,
  Karl~Moritz Hermann, G{\'a}bor Melis, and Edward Grefenstette. 2018.
\newblock \href {https://doi.org/10.1162/tacl_a_00023} {The {N}arrative{QA}
  reading comprehension challenge}.
\newblock \emph{Transactions of the Association for Computational Linguistics},
  6:317--328.

\bibitem[{Kumar et~al.(2020)Kumar, Choudhary, and Cho}]{kumar-etal-2020-data}
Varun Kumar, Ashutosh Choudhary, and Eunah Cho. 2020.
\newblock \href {https://aclanthology.org/2020.lifelongnlp-1.3} {Data
  augmentation using pre-trained transformer models}.
\newblock In \emph{Proceedings of the 2nd Workshop on Life-long Learning for
  Spoken Language Systems}, pages 18--26, Suzhou, China. Association for
  Computational Linguistics.

\bibitem[{Lewis et~al.(2020)Lewis, Liu, Goyal, Ghazvininejad, Mohamed, Levy,
  Stoyanov, and Zettlemoyer}]{lewis-etal-2020-bart}
Mike Lewis, Yinhan Liu, Naman Goyal, Marjan Ghazvininejad, Abdelrahman Mohamed,
  Omer Levy, Veselin Stoyanov, and Luke Zettlemoyer. 2020.
\newblock \href {https://doi.org/10.18653/v1/2020.acl-main.703} {{BART}:
  Denoising sequence-to-sequence pre-training for natural language generation,
  translation, and comprehension}.
\newblock In \emph{Proceedings of the 58th Annual Meeting of the Association
  for Computational Linguistics}, pages 7871--7880, Online. Association for
  Computational Linguistics.

\bibitem[{Lin et~al.(2021)Lin, Mihaylov, Artetxe, Wang, Chen, Simig, Ott,
  Goyal, Bhosale, Du et~al.}]{lin2021few}
Xi~Victoria Lin, Todor Mihaylov, Mikel Artetxe, Tianlu Wang, Shuohui Chen,
  Daniel Simig, Myle Ott, Naman Goyal, Shruti Bhosale, Jingfei Du, et~al. 2021.
\newblock Few-shot learning with multilingual language models.
\newblock \emph{arXiv preprint arXiv:2112.10668}.

\bibitem[{Maas et~al.(2011)Maas, Daly, Pham, Huang, Ng, and
  Potts}]{maas-EtAl:2011:ACL-HLT2011}
Andrew~L. Maas, Raymond~E. Daly, Peter~T. Pham, Dan Huang, Andrew~Y. Ng, and
  Christopher Potts. 2011.
\newblock \href {http://www.aclweb.org/anthology/P11-1015} {Learning word
  vectors for sentiment analysis}.
\newblock In \emph{Proceedings of the 49th Annual Meeting of the Association
  for Computational Linguistics: Human Language Technologies}, pages 142--150,
  Portland, Oregon, USA. Association for Computational Linguistics.

\bibitem[{McCann et~al.(2018)McCann, Keskar, Xiong, and
  Socher}]{McCann2018decaNLP}
Bryan McCann, Nitish~Shirish Keskar, Caiming Xiong, and Richard Socher. 2018.
\newblock The natural language decathlon: Multitask learning as question
  answering.
\newblock \emph{arXiv preprint arXiv:1806.08730}.

\bibitem[{Mekala et~al.(2021)Mekala, Gangal, and
  Shang}]{mekala-etal-2021-coarse2fine}
Dheeraj Mekala, Varun Gangal, and Jingbo Shang. 2021.
\newblock \href {https://doi.org/10.18653/v1/2021.emnlp-main.46}
  {{C}oarse2{F}ine: Fine-grained text classification on coarsely-grained
  annotated data}.
\newblock In \emph{Proceedings of the 2021 Conference on Empirical Methods in
  Natural Language Processing}, pages 583--594, Online and Punta Cana,
  Dominican Republic. Association for Computational Linguistics.

\bibitem[{Mekala and Shang(2020)}]{mekala2020contextualized}
Dheeraj Mekala and Jingbo Shang. 2020.
\newblock Contextualized weak supervision for text classification.
\newblock In \emph{Proceedings of the 58th Annual Meeting of the Association
  for Computational Linguistics}, pages 323--333.

\bibitem[{Mekala et~al.(2020)Mekala, Zhang, and Shang}]{mekala2020meta}
Dheeraj Mekala, Xinyang Zhang, and Jingbo Shang. 2020.
\newblock Meta: Metadata-empowered weak supervision for text classification.
\newblock In \emph{Proceedings of the 2020 Conference on Empirical Methods in
  Natural Language Processing (EMNLP)}.

\bibitem[{Oliver et~al.(2018)Oliver, Odena, Raffel, Cubuk, and
  Goodfellow}]{NEURIPS2018_c1fea270}
Avital Oliver, Augustus Odena, Colin~A Raffel, Ekin~Dogus Cubuk, and Ian
  Goodfellow. 2018.
\newblock \href
  {https://proceedings.neurips.cc/paper/2018/file/c1fea270c48e8079d8ddf7d06d26ab52-Paper.pdf}
  {Realistic evaluation of deep semi-supervised learning algorithms}.
\newblock In \emph{Advances in Neural Information Processing Systems},
  volume~31. Curran Associates, Inc.

\bibitem[{Phang et~al.(2018)Phang, F{\'e}vry, and Bowman}]{Phang2018SentenceEO}
Jason Phang, Thibault F{\'e}vry, and Samuel~R. Bowman. 2018.
\newblock Sentence encoders on stilts: Supplementary training on intermediate
  labeled-data tasks.
\newblock \emph{ArXiv}, abs/1811.01088.

\bibitem[{Pruksachatkun et~al.(2020)Pruksachatkun, Phang, Liu, Htut, Zhang,
  Pang, Vania, Kann, and Bowman}]{pruksachatkun-etal-2020-intermediate}
Yada Pruksachatkun, Jason Phang, Haokun Liu, Phu~Mon Htut, Xiaoyi Zhang,
  Richard~Yuanzhe Pang, Clara Vania, Katharina Kann, and Samuel~R. Bowman.
  2020.
\newblock \href {https://doi.org/10.18653/v1/2020.acl-main.467}
  {Intermediate-task transfer learning with pretrained language models: When
  and why does it work?}
\newblock In \emph{Proceedings of the 58th Annual Meeting of the Association
  for Computational Linguistics}, pages 5231--5247, Online. Association for
  Computational Linguistics.

\bibitem[{Pryzant et~al.(2022)Pryzant, Yang, Xu, Zhu, and
  Zeng}]{Pryzant2022AutomaticRI}
Reid Pryzant, Ziyi Yang, Yichong Xu, Chenguang Zhu, and Michael Zeng. 2022.
\newblock Automatic rule induction for efficient semi-supervised learning.
\newblock In \emph{Arxiv}.

\bibitem[{Puri et~al.(2020)Puri, Spring, Shoeybi, Patwary, and
  Catanzaro}]{puri-etal-2020-training}
Raul Puri, Ryan Spring, Mohammad Shoeybi, Mostofa Patwary, and Bryan Catanzaro.
  2020.
\newblock \href {https://doi.org/10.18653/v1/2020.emnlp-main.468} {Training
  question answering models from synthetic data}.
\newblock In \emph{Proceedings of the 2020 Conference on Empirical Methods in
  Natural Language Processing (EMNLP)}, pages 5811--5826, Online. Association
  for Computational Linguistics.

\bibitem[{Radford and Narasimhan(2018)}]{Radford2018ImprovingLU}
Alec Radford and Karthik Narasimhan. 2018.
\newblock Improving language understanding by generative pre-training.
\newblock In \emph{Arxiv}.

\bibitem[{Radford et~al.(2019)Radford, Wu, Child, Luan, Amodei, and
  Sutskever}]{Radford2019LanguageMA}
Alec Radford, Jeff Wu, Rewon Child, David Luan, Dario Amodei, and Ilya
  Sutskever. 2019.
\newblock Language models are unsupervised multitask learners.
\newblock In \emph{Arxiv}.

\bibitem[{Raffel et~al.(2020)Raffel, Shazeer, Roberts, Lee, Narang, Matena,
  Zhou, Li, and Liu}]{2020t5}
Colin Raffel, Noam Shazeer, Adam Roberts, Katherine Lee, Sharan Narang, Michael
  Matena, Yanqi Zhou, Wei Li, and Peter~J. Liu. 2020.
\newblock \href {http://jmlr.org/papers/v21/20-074.html} {Exploring the limits
  of transfer learning with a unified text-to-text transformer}.
\newblock \emph{Journal of Machine Learning Research}, 21(140):1--67.

\bibitem[{Rajpurkar et~al.(2018)Rajpurkar, Jia, and
  Liang}]{rajpurkar-etal-2018-know}
Pranav Rajpurkar, Robin Jia, and Percy Liang. 2018.
\newblock \href {https://doi.org/10.18653/v1/P18-2124} {Know what you don{'}t
  know: Unanswerable questions for {SQ}u{AD}}.
\newblock In \emph{Proceedings of the 56th Annual Meeting of the Association
  for Computational Linguistics (Volume 2: Short Papers)}, pages 784--789,
  Melbourne, Australia. Association for Computational Linguistics.

\bibitem[{Rajpurkar et~al.(2016)Rajpurkar, Zhang, Lopyrev, and
  Liang}]{rajpurkar-etal-2016-squad}
Pranav Rajpurkar, Jian Zhang, Konstantin Lopyrev, and Percy Liang. 2016.
\newblock \href {https://doi.org/10.18653/v1/D16-1264} {{SQ}u{AD}: 100,000+
  questions for machine comprehension of text}.
\newblock In \emph{Proceedings of the 2016 Conference on Empirical Methods in
  Natural Language Processing}, pages 2383--2392, Austin, Texas. Association
  for Computational Linguistics.

\bibitem[{Reddy et~al.(2019)Reddy, Chen, and Manning}]{reddy2019coqa}
Siva Reddy, Danqi Chen, and Christopher~D Manning. 2019.
\newblock Coqa: A conversational question answering challenge.
\newblock \emph{Transactions of the Association for Computational Linguistics},
  7:249--266.

\bibitem[{Sap et~al.(2019)Sap, Rashkin, Chen, Le~Bras, and
  Choi}]{sap-etal-2019-social}
Maarten Sap, Hannah Rashkin, Derek Chen, Ronan Le~Bras, and Yejin Choi. 2019.
\newblock \href {https://doi.org/10.18653/v1/D19-1454} {Social {IQ}a:
  Commonsense reasoning about social interactions}.
\newblock In \emph{Proceedings of the 2019 Conference on Empirical Methods in
  Natural Language Processing and the 9th International Joint Conference on
  Natural Language Processing (EMNLP-IJCNLP)}, pages 4463--4473, Hong Kong,
  China. Association for Computational Linguistics.

\bibitem[{Schick and
  Sch{\"u}tze(2021{\natexlab{a}})}]{schick-schutze-2021-exploiting}
Timo Schick and Hinrich Sch{\"u}tze. 2021{\natexlab{a}}.
\newblock \href {https://doi.org/10.18653/v1/2021.eacl-main.20} {Exploiting
  cloze-questions for few-shot text classification and natural language
  inference}.
\newblock In \emph{Proceedings of the 16th Conference of the European Chapter
  of the Association for Computational Linguistics: Main Volume}, pages
  255--269, Online. Association for Computational Linguistics.

\bibitem[{Schick and
  Sch{\"u}tze(2021{\natexlab{b}})}]{schick-schutze-2021-generating}
Timo Schick and Hinrich Sch{\"u}tze. 2021{\natexlab{b}}.
\newblock \href {https://doi.org/10.18653/v1/2021.emnlp-main.555} {Generating
  datasets with pretrained language models}.
\newblock In \emph{Proceedings of the 2021 Conference on Empirical Methods in
  Natural Language Processing}, pages 6943--6951, Online and Punta Cana,
  Dominican Republic. Association for Computational Linguistics.

\bibitem[{Sennrich et~al.(2016)Sennrich, Haddow, and
  Birch}]{sennrich-etal-2016-improving}
Rico Sennrich, Barry Haddow, and Alexandra Birch. 2016.
\newblock \href {https://doi.org/10.18653/v1/P16-1009} {Improving neural
  machine translation models with monolingual data}.
\newblock In \emph{Proceedings of the 54th Annual Meeting of the Association
  for Computational Linguistics (Volume 1: Long Papers)}, pages 86--96, Berlin,
  Germany. Association for Computational Linguistics.

\bibitem[{Socher et~al.(2013)Socher, Perelygin, Wu, Chuang, Manning, Ng, and
  Potts}]{socher-etal-2013-recursive}
Richard Socher, Alex Perelygin, Jean Wu, Jason Chuang, Christopher~D. Manning,
  Andrew Ng, and Christopher Potts. 2013.
\newblock \href {https://aclanthology.org/D13-1170} {Recursive deep models for
  semantic compositionality over a sentiment treebank}.
\newblock In \emph{Proceedings of the 2013 Conference on Empirical Methods in
  Natural Language Processing}, pages 1631--1642, Seattle, Washington, USA.
  Association for Computational Linguistics.

\bibitem[{Tam et~al.(2021)Tam, R.~Menon, Bansal, Srivastava, and
  Raffel}]{tam-etal-2021-improving}
Derek Tam, Rakesh R.~Menon, Mohit Bansal, Shashank Srivastava, and Colin
  Raffel. 2021.
\newblock \href {https://doi.org/10.18653/v1/2021.emnlp-main.407} {Improving
  and simplifying pattern exploiting training}.
\newblock In \emph{Proceedings of the 2021 Conference on Empirical Methods in
  Natural Language Processing}, pages 4980--4991, Online and Punta Cana,
  Dominican Republic. Association for Computational Linguistics.

\bibitem[{Trischler et~al.(2017)Trischler, Wang, Yuan, Harris, Sordoni,
  Bachman, and Suleman}]{trischler-etal-2017-newsqa}
Adam Trischler, Tong Wang, Xingdi Yuan, Justin Harris, Alessandro Sordoni,
  Philip Bachman, and Kaheer Suleman. 2017.
\newblock \href {https://doi.org/10.18653/v1/W17-2623} {{N}ews{QA}: A machine
  comprehension dataset}.
\newblock In \emph{Proceedings of the 2nd Workshop on Representation Learning
  for {NLP}}, pages 191--200, Vancouver, Canada. Association for Computational
  Linguistics.

\bibitem[{Vu et~al.(2022{\natexlab{a}})Vu, Barua, Lester, Cer, Iyyer, and
  Constant}]{vu-etal-2022-overcoming}
Tu~Vu, Aditya Barua, Brian Lester, Daniel Cer, Mohit Iyyer, and Noah Constant.
  2022{\natexlab{a}}.
\newblock \href {https://arxiv.org/abs/2205.12647} {Overcoming catastrophic
  forgetting in zero-shot cross-lingual generation}.
\newblock \emph{arXiv preprint arXiv:2205.12647}.

\bibitem[{Vu et~al.(2022{\natexlab{b}})Vu, Lester, Constant, Al-Rfou{'}, and
  Cer}]{vu-etal-2022-spot}
Tu~Vu, Brian Lester, Noah Constant, Rami Al-Rfou{'}, and Daniel Cer.
  2022{\natexlab{b}}.
\newblock \href {https://doi.org/10.18653/v1/2022.acl-long.346} {{SP}o{T}:
  Better frozen model adaptation through soft prompt transfer}.
\newblock In \emph{Proceedings of the 60th Annual Meeting of the Association
  for Computational Linguistics (Volume 1: Long Papers)}, pages 5039--5059,
  Dublin, Ireland. Association for Computational Linguistics.

\bibitem[{Vu et~al.(2021)Vu, Luong, Le, Simon, and Iyyer}]{vu-etal-2021-strata}
Tu~Vu, Minh-Thang Luong, Quoc Le, Grady Simon, and Mohit Iyyer. 2021.
\newblock \href {https://doi.org/10.18653/v1/2021.emnlp-main.462} {{ST}ra{TA}:
  Self-training with task augmentation for better few-shot learning}.
\newblock In \emph{Proceedings of the 2021 Conference on Empirical Methods in
  Natural Language Processing}, pages 5715--5731, Online and Punta Cana,
  Dominican Republic. Association for Computational Linguistics.

\bibitem[{Vu et~al.(2020)Vu, Wang, Munkhdalai, Sordoni, Trischler,
  Mattarella-Micke, Maji, and Iyyer}]{vu-etal-2020-exploring}
Tu~Vu, Tong Wang, Tsendsuren Munkhdalai, Alessandro Sordoni, Adam Trischler,
  Andrew Mattarella-Micke, Subhransu Maji, and Mohit Iyyer. 2020.
\newblock \href {https://doi.org/10.18653/v1/2020.emnlp-main.635} {Exploring
  and predicting transferability across {NLP} tasks}.
\newblock In \emph{Proceedings of the 2020 Conference on Empirical Methods in
  Natural Language Processing (EMNLP)}, pages 7882--7926, Online. Association
  for Computational Linguistics.

\bibitem[{Wei and Zou(2019)}]{wei-zou-2019-eda}
Jason Wei and Kai Zou. 2019.
\newblock \href {https://doi.org/10.18653/v1/D19-1670} {{EDA}: Easy data
  augmentation techniques for boosting performance on text classification
  tasks}.
\newblock In \emph{Proceedings of the 2019 Conference on Empirical Methods in
  Natural Language Processing and the 9th International Joint Conference on
  Natural Language Processing (EMNLP-IJCNLP)}, pages 6382--6388, Hong Kong,
  China. Association for Computational Linguistics.

\bibitem[{Williams et~al.(2018)Williams, Nangia, and Bowman}]{N18-1101}
Adina Williams, Nikita Nangia, and Samuel Bowman. 2018.
\newblock \href {http://aclweb.org/anthology/N18-1101} {A broad-coverage
  challenge corpus for sentence understanding through inference}.
\newblock In \emph{Proceedings of the 2018 Conference of the North American
  Chapter of the Association for Computational Linguistics: Human Language
  Technologies, Volume 1 (Long Papers)}, pages 1112--1122. Association for
  Computational Linguistics.

\bibitem[{Wolf et~al.(2020)Wolf, Debut, Sanh, Chaumond, Delangue, Moi, Cistac,
  Rault, Louf, Funtowicz, Davison, Shleifer, von Platen, Ma, Jernite, Plu, Xu,
  Le~Scao, Gugger, Drame, Lhoest, and Rush}]{wolf-etal-2020-transformers}
Thomas Wolf, Lysandre Debut, Victor Sanh, Julien Chaumond, Clement Delangue,
  Anthony Moi, Pierric Cistac, Tim Rault, Remi Louf, Morgan Funtowicz, Joe
  Davison, Sam Shleifer, Patrick von Platen, Clara Ma, Yacine Jernite, Julien
  Plu, Canwen Xu, Teven Le~Scao, Sylvain Gugger, Mariama Drame, Quentin Lhoest,
  and Alexander Rush. 2020.
\newblock \href {https://doi.org/10.18653/v1/2020.emnlp-demos.6} {Transformers:
  State-of-the-art natural language processing}.
\newblock In \emph{Proceedings of the 2020 Conference on Empirical Methods in
  Natural Language Processing: System Demonstrations}, pages 38--45, Online.
  Association for Computational Linguistics.

\bibitem[{Xiong et~al.(2019)Xiong, Wu, Wang, Kulkarni, Yu, Guo, Chang, and
  Wang}]{xiong2019tweetqa}
Wenhan Xiong, Jiawei Wu, Hong Wang, Vivek Kulkarni, Mo~Yu, Xiaoxiao Guo, Shiyu
  Chang, and William~Yang Wang. 2019.
\newblock Tweetqa: A social media focused question answering dataset.
\newblock In \emph{Proceedings of the 57th Annual Meeting of the Association
  for Computational Linguistics}.

\bibitem[{Yang et~al.(2020)Yang, Malaviya, Fernandez, Swayamdipta, Le~Bras,
  Wang, Bhagavatula, Choi, and Downey}]{yang-etal-2020-generative}
Yiben Yang, Chaitanya Malaviya, Jared Fernandez, Swabha Swayamdipta, Ronan
  Le~Bras, Ji-Ping Wang, Chandra Bhagavatula, Yejin Choi, and Doug Downey.
  2020.
\newblock \href {https://doi.org/10.18653/v1/2020.findings-emnlp.90}
  {Generative data augmentation for commonsense reasoning}.
\newblock In \emph{Findings of the Association for Computational Linguistics:
  EMNLP 2020}, pages 1008--1025, Online. Association for Computational
  Linguistics.

\bibitem[{Yogatama et~al.(2019)Yogatama, de~Masson~d'Autume, Connor,
  Kocisk{\'y}, Chrzanowski, Kong, Lazaridou, Ling, Yu, Dyer, and
  Blunsom}]{Yogatama2019LearningAE}
Dani Yogatama, Cyprien de~Masson~d'Autume, Jerome~T. Connor, Tom{\'a}s
  Kocisk{\'y}, Mike Chrzanowski, Lingpeng Kong, Angeliki Lazaridou, Wang Ling,
  Lei Yu, Chris Dyer, and Phil Blunsom. 2019.
\newblock Learning and evaluating general linguistic intelligence.
\newblock \emph{ArXiv}, abs/1901.11373.

\bibitem[{Zhang et~al.(2019)Zhang, Zhao, Saleh, and Liu}]{zhang2019pegasus}
Jingqing Zhang, Yao Zhao, Mohammad Saleh, and Peter~J. Liu. 2019.
\newblock \href {http://arxiv.org/abs/1912.08777} {Pegasus: Pre-training with
  extracted gap-sentences for abstractive summarization}.

\bibitem[{Zhang et~al.(2015)Zhang, Zhao, and LeCun}]{zhang2015characterlevel}
Xiang Zhang, Junbo Zhao, and Yann LeCun. 2015.
\newblock \href
  {http://papers.nips.cc/paper/5782-character-level-convolutional-networks-for-text-classification.pdf}
  {Character-level convolutional networks for text classification}.
\newblock In C.~Cortes, N.~D. Lawrence, D.~D. Lee, M.~Sugiyama, and R.~Garnett,
  editors, \emph{Advances in Neural Information Processing Systems 28}, pages
  649--657. Curran Associates, Inc.

\end{thebibliography}
\bibliographystyle{acl_natbib}

\newpage
\appendix

\newpage
\section{Appendix}
\label{sec:appendix}

\subsection{Target Task Datasets}
\label{app:target_task_datas}
The details of target task datasets are as follows:
\begin{itemize}[leftmargin=*,nosep]
    \item \textbf{IMDb:}~\cite{maas-EtAl:2011:ACL-HLT2011} is a movie review dataset with positive and negative as sentiments.
    \item \textbf{Yelp:}\footnote{\url{https://www.yelp.com/dataset/}} is a collection of reviews written by Yelp users with five fine-grained sentiment ratings.
    \item \textbf{SST-2:}~\cite{socher-etal-2013-recursive} is a binary sentiment classification dataset with single sentence texts.
    \item \textbf{Yahoo:}~\cite{zhang2015characterlevel} is a topic classification dataset with question and answer pairs. Using these pairs, the task is to predict their corresponding topic.
    \item \textbf{The New York Times (NYT):} : contains news articles written and published by The New York Times that are classified into 5 wide genres.
    \item \textbf{AGNews:}~\cite{zhang2015characterlevel} is a topic categorization dataset in news domain from AG’s corpus.
\end{itemize}
The size of test sets is mentioned in Table~\ref{tbl:test_datastats}.

\begin{table}[t]
    \center
    \caption{Dataset statistics.}
    \label{tbl:test_datastats}
    \vspace{-3mm}
    % \small
    \begin{tabular}{c c c c}
        \toprule
        {\textbf{Dataset}} & {\textbf{\# Test Examples}} \\
        \midrule
        \textbf{IMDb} & 25000 \\
        \textbf{Yelp} & 50000 \\
        \textbf{SST-2} & 2211 \\
        \textbf{Yahoo} & 60000\\
        \textbf{NYT} & 10582\\
        \textbf{AGNews} & 114000\\
        \bottomrule
    \end{tabular}
\end{table}

\begin{table*}[t]
    \center
    \caption{Evaluation Results with validation set.}
    \vspace{-3mm}
    \scalebox{0.6}{
    % \small
    \begin{tabular}{c cccccc cccccc}
        \toprule
                         & \multicolumn{6}{c}{\textbf{Sentiment}} & \multicolumn{6}{c}{\textbf{Topic}} \\
        % \cmidrule{2-13}
        \cmidrule(lr){2-7} \cmidrule(lr){8-13}
                         & \multicolumn{2}{c}{\textbf{IMDb}} & \multicolumn{2}{c}{\textbf{Yelp}} & \multicolumn{2}{c}{\textbf{SST-2}} &  \multicolumn{2}{c}{\textbf{NYT}} & \multicolumn{2}{c}{\textbf{Yahoo}} & \multicolumn{2}{c}{\textbf{AGNews}}
                         \\
        \textbf{Method} & Mi-F$_1$ & Ma-F$_1$ & Mi-F$_1$ & Ma-F$_1$ & Mi-F$_1$ & Ma-F$_1$ & Mi-F$_1$ & Ma-F$_1$ & Mi-F$_1$ & Ma-F$_1$ & Mi-F$_1$ & Ma-F$_1$ \\
        \midrule
        BERT-FT & $68.7_{3.6}$ & $68.5_{3.6}$ & $38.1_{5.2}$ & $36.3_{5.2}$ & $57.5_{1.4}$ & $55.8_{2.3}$ & $88.7_{5.5}$ & $83.9_{4.6}$ & $54.4_{1.8}$ & $53.9_{1.2}$ & $74.6_{5.6}$ & $74.7_{5.3}$ \\
        ITFT-MNLI & $66.2_{3.3}$ & $64.6_{4.2}$ & $35.3_{3.7}$ & $33.3_{3.5}$ & $60.8_{1.8}$ & $58.7{1.6}$ & $78.0_{2.9}$ & $63.1_{6.2}$ & $28.2_{4.5}$ & $27.2_{4.2}$ & $52.9_{3.2}$ & $51.4_{4.5}$ \\
        ITFT-SQuAD & $61.1_{2.0}$ & $59.7_{2.6}$ & $34.0_{2.3}$ & $31.6_{3.6}$ & $56.5_{1.4}$ & $56.0_{1.5}$ & $88.9_{2.2}$ & $75.8_{4.6}$ & $36.2_{4.0}$ & $35.3_{4.4}$ & $58.2_{5.5}$ & $56.2_{6.6}$\\
        BackTranslation & $67.4_{2.0}$ & $66.9_{2.0}$ & \underline{$38.7_{3.6}$} & \underline{$36.4_{4.4}$} & $61.0_{4.3}$ & $60.4_{5.0}$ & $93.7_{1.4}$ & $88.7_{0.3}$ & \underline{$56.8_{1.4}$} & $56.2_{1.2}$ & \underline{$80.3_{2.0}$} & \underline{$80.3_{2.0}$} \\
        PEGASUS & $66.2_{3.8}$ & $65.3_{3.9}$ & $32.8_{7.0}$ & $29.5_{8.0}$ & \underline{$61.9_{3.9}$} & \underline{$60.6_{3.9}$} & $93.9_{0.6}$ & $87.3_{1.7}$ & $57.7_{3.0}$ & \underline{$56.3_{1.0}$} & $79.7_{1.6}$ & $79.9_{1.6}$\\
        EDA & $63.3_{4.2}$ & $61.6_{4.4}$ & $32.5_{6.7}$ & $30.6_{8.2}$ & $58.8_{2.9}$ & $58.1_{3.4}$ & \underline{$95.7_{0.7}$} & \underline{$90.6_{2.0}$} & $55.7_{1.1}$ & $56.3_{1.0}$ & $79.8_{0.7}$ & $79.9_{0.4}$\\
        $\our\backslash QA$ & \underline{$71.8_{4.5}$} & \underline{$71.1_{4.9}$} & $38.0_{0.3}$ & $36.0_{0.5}$ & $60.1_{4.7}$ & $58.0_{6.2}$ & $92.3_{0.5}$ & $84.2_{0.5}$ & $53.8_{0.9}$ & $52.8_{0.7}$ & $80.0_{1.6}$ & $79.6_{1.7}$ \\
        \midrule
        \our-SQuAD & $58.5_{2.5}$ & $56.5_{1.7}$ & $37.6_{1.6}$ & $36.4_{0.5}$ & $56.3_{2.2}$ & $55.8_{1.9}$ & $93.4_{0.5}$ & $86.6_{0.9}$ & $56.1_{1.7}$ & $54.8_{1.8}$ & $\mathbf{82.1_{0.2}}$ & $\mathbf{82.1_{0.2}}$ \\
        \our-NewsQA & $61.5_{6.7}$ & $60.1_{8.1}$ & $34.8_{0.9}$ & $32.3_{2.5}$ & $57.1_{5.8}$ & $56.2_{6.2}$ & $92.5_{0.8}$ & $83.8_{1.5}$ & $55.3_{2.5}$ & $54.8_{2.8}$ & $\mathbf{80.6_{3.7}}$ & $80.3_{3.9}$ \\
        \our-TweetQA & $\mathbf{78.3_{2.8}}$ & $\mathbf{78.1_{3.0}}$ & $\mathbf{41.3_{0.5}}$ & $\mathbf{37.7_{3.5}}$ & $\mathbf{71.6_{4.8}}$ & $\mathbf{70.9_{5.3}}$ & $93.1_{1.5}$ & $85.5_{3.0}$ & $\mathbf{58.8_{1.6}}$ & $\mathbf{57.9_{2.4}}$ & $\mathbf{81.2_{1.9}}$ & $\mathbf{81.1_{1.9}}$  \\
        \our-SocialIQA & $\mathbf{78.3_{2.2}}$ & $\mathbf{78.1_{1.5}}$ & $\mathbf{41.5_{0.8}}$ & $\mathbf{39.0_{1.8}}$ & $\mathbf{74.4_{3.9}}$ & $\mathbf{74.3_{4.0}}$ & $92.3_{0.7}$ & $84.7_{1.3}$ & $\mathbf{58.5_{2.0}}$ & $\mathbf{58.0_{2.3}}$ & $\mathbf{82.4_{1.6}}$ & $\mathbf{82.2_{1.6}}$\\
        \our-CosmosQA & $\mathbf{74.1_{4.9}}$ & $\mathbf{73.6_{5.5}}$ & $\mathbf{38.9_{2.1}}$ & $31.3_{4.3}$ & $\mathbf{64.5_{2.3}}$ & $\mathbf{63.3_{3.1}}$ & $93.6_{1.1}$ & $86.2_{2.2}$ & $\mathbf{59.4_{0.6}}$ & $\mathbf{59.3_{0.1}}$ & $\mathbf{82.7_{1.3}}$ & $\mathbf{82.5_{1.4}}$ \\
        \bottomrule
    \end{tabular}
    }
    \vspace{-3mm}
    \label{tbl:f1_val_results}
\end{table*}

\subsection{Performance vs k}
\label{app:perf_topk}
\begin{figure}[t]
    \includegraphics[width=0.9\linewidth]{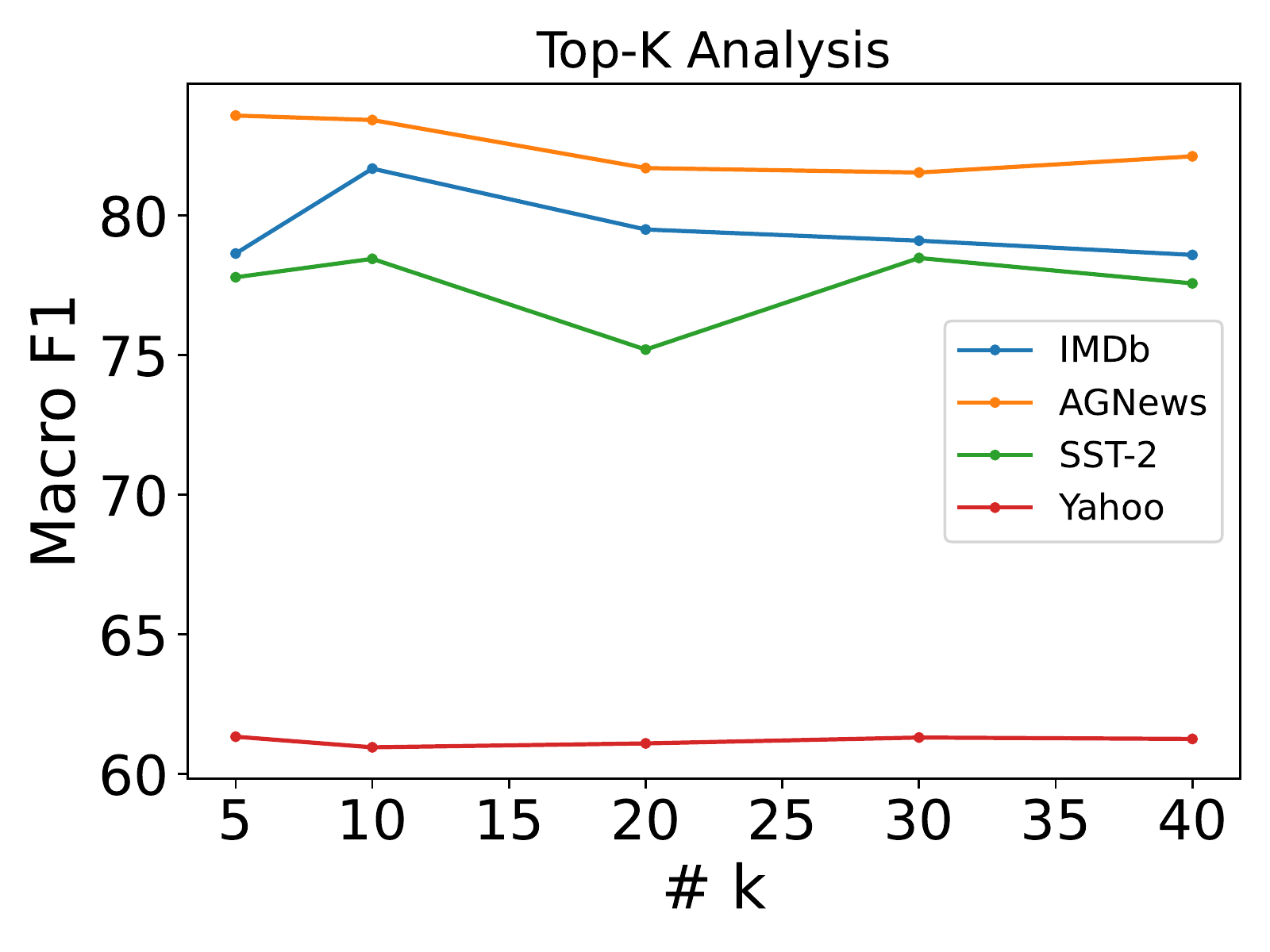}
    \vspace{-3mm}
    \caption{Macro-F$_1$ scores of \our-SocialIQA w.r.t. k. Each experiment is repeated with three different seeds and mean performance is plotted.}
    \label{figure:perf_k}
    \vspace{-5mm}
\end{figure}
We vary k in top-k sampling and plot the performance of \our-SocialIQA on IMDb, SST-2, AGNews, and Yahoo datasets in Figure~\ref{figure:perf_k}.
We fix the few-shot supervision size to 8 samples per label and generate 450 samples per label. We repeat each experiment thrice and plot the mean performance.
Upon manual inspection, We observe that the samples generated with k=20 are more diverse than k=10, however, the influence of k on performance is not significant.

\subsection{Experiments with a validation set}
We perform experiments with a validation set.
Since large validation sets are impractical in few-shot settings~\cite{NEURIPS2018_c1fea270}, we consider the validation set to be of same size as the few-shot training set i.e. 8 annotated samples per label.
In the experiments with validation set, we perform early stopping based on validation set performance.
We present experimental results on few-shot setting with validation set in Table~\ref{tbl:f1_val_results}.
We seldom observe significant improvement upon introducing the validation set. 
This is because a small validation set which is of same size as few-shot supervision is not large enough to tune the hyperparameters.

\begin{table*}[t]
    \center
    % \caption{Dataset statistics. \jingbo{Make it two-column and add all your final label names into the table?}}
    \vspace{-3mm}
    \caption{Example generated texts using our method with all QA datasets}
    \label{tbl:analysis_gen}
    \small
     \scalebox{1}{
    \begin{tabular}{c c c c c}
        \toprule
            {\textbf{Dataset}} & {\textbf{Label}} & {\textbf{QA Dataset}} & {\textbf{Generated Text}}  \\
        \midrule
        \multirow{11}{*}{IMDb} & \multirow{11}{*}{good} & SQuAD &  \begin{tabular}{@{}c@{}} \scriptsize the film was first screened in the united kingdom and ireland in june 1985,                                               \\ \scriptsize where it received generally favourable reviews. \end{tabular}\\
        \cmidrule{3-4}
                                                      & & NewsQA & \begin{tabular}{@{}c@{}} \scriptsize i love that "babylon 5" has a story in the form of an \\ \scriptsize animated television show. when the first "babylon" appeared on the vhs channel, i was a kid. \end{tabular}\\
        \cmidrule{3-4}
                                                      & & TweetQA & \begin{tabular}{@{}c@{}} \scriptsize this is a fantastic film, and i loved the trailer. you may know me \\ \scriptsize and you may also know a friend of mine,  so i can totally see \end{tabular}\\
        \cmidrule{3-4}
                                                      & & SocialIQA & \begin{tabular}{@{}c@{}} \scriptsize skylar went to the movies and saw a good movie. \end{tabular}\\
        \cmidrule{3-4}                      
                                                      & & CosmosQA & \begin{tabular}{@{}c@{}} \scriptsize this is one of my favorite movies of all time because the movie was made way back \\ \scriptsize in the 80s and the dialogue is pretty hilarious. \end{tabular}\\
        \midrule
        \multirow{11}{*}{AGNews} & \multirow{11}{*}{sports} & SQuAD &  \begin{tabular}{@{}c@{}} \scriptsize this document is about sports, and details the general aspects                                                                                      of the  sport. it lays out the sport's rules, \\ \scriptsize  and sets out the objectives                                                                                      and common grounds upon which the sport is based. \end{tabular}\\
        \cmidrule{3-4}
                                                      & & NewsQA & \begin{tabular}{@{}c@{}} \scriptsize (cnn) -- when the cincinnati bengals and new england patriots \\ \scriptsize played at a stadium on friday night, an audience of about \\ \scriptsize 2 million people was in tow. and those people weren't the only ones enjoying the game. \end{tabular}\\
        \cmidrule{3-4}
                                                      & & TweetQA & \begin{tabular}{@{}c@{}} \scriptsize it's a football document and i want to make sure all our fans know. \#fcbtampa. \\ \scriptsize thanks for the \#baltimorebluhnnnnnnnnnt \end{tabular}\\
        \cmidrule{3-4}
                                                      & & SocialIQA & \begin{tabular}{@{}c@{}} \scriptsize carson got the ball in their possession after scoring a goal in the soccer match. \end{tabular}\\
        \cmidrule{3-4}                      
                                                      & & CosmosQA & \begin{tabular}{@{}c@{}} \scriptsize we just played a nice game, and i thought we 'd be better off if they could \\ \scriptsize come up with a good plan to do it. we did have a little time, \\ \scriptsize however, and we 'd have a chance to score and give the ball back, but it just never worked out \end{tabular}\\
        \bottomrule
    \end{tabular}
    }
\end{table*}

\subsection{Examples of Generated Training Data}
Table~\ref{tbl:analysis_gen} shows a few examples of synthetic training data corresponding to IMDb and AGNews datasets generated by our method with all QA datasets.

\end{document}